\documentclass[twoside,11pt]{article}

\usepackage[abbrvbib, preprint]{jmlr2e}

\usepackage{lastpage}
\jmlrheading{}{2025}{1-\pageref{LastPage}}{10/25}{}{}{Youngjae Min and Navid Azizan}

\ShortHeadings{HardNet: Hard-Constrained Neural Networks}{Min and Azizan}
\firstpageno{1}

\usepackage[utf8]{inputenc} 
\usepackage[T1]{fontenc}    
\usepackage{hyperref}       
\usepackage{url}            
\usepackage{booktabs}       
\usepackage{amsfonts}       
\usepackage{nicefrac}       
\usepackage{microtype}      
\usepackage[table]{xcolor}         
\usepackage{comment}

\title{HardNet: Hard-Constrained Neural Networks\\with Universal Approximation Guarantees}

\author{\name Youngjae Min \email yjm@mit.edu \\
       \addr 
       Massachusetts Institute of Technology
       \AND
       \name Navid Azizan \email azizan@mit.edu \\
       \addr 
       Massachusetts Institute of Technology}

\editor{My editor}

\usepackage{amsmath}
\usepackage{mathtools}
\usepackage{thmtools}
\usepackage{thm-restate}

\newtheorem{proposition}[theorem]{Proposition}

\newtheorem{assumption}[theorem]{Assumption}

\DeclareMathOperator*{\argmin}{arg\,min}

\newcommand{\method}{\mbox{\textup{\textsf{HardNet}}}}
\newcommand{\methodaff}{\mbox{\textup{\textsf{HardNet-Aff}}}}
\newcommand{\methodcvx}{\mbox{\textup{\textsf{HardNet-Cvx}}}}
\newcommand{\relu}{\mathsf{ReLU}}

\usepackage{enumitem}
\usepackage{wrapfig}
\usepackage{arydshln}
\usepackage{dsfont}

\begin{document}

\maketitle

\begin{abstract}%
  Incorporating prior knowledge or specifications of input-output relationships into machine learning models has attracted significant attention, as it enhances generalization from limited data and yields conforming outputs. However, most existing approaches use soft constraints by penalizing violations through regularization, which offers no guarantee of constraint satisfaction, especially on inputs far from the training distribution---an essential requirement in safety-critical applications. On the other hand, imposing hard constraints on neural networks may hinder their representational power, adversely affecting performance. To address this, we propose \method{}, a practical framework for constructing neural networks that inherently satisfy hard constraints without sacrificing model capacity.
  Unlike approaches that modify outputs only at inference time, \method{} enables end-to-end training with hard constraint guarantees, leading to improved performance.
  To the best of our knowledge, \method{} is the first method
  that enables efficient and differentiable enforcement of
  more than one \emph{input-dependent} inequality constraint. It allows unconstrained optimization of the network parameters using standard algorithms by appending a differentiable \emph{closed-form} enforcement layer to the network's output. Furthermore, we show that \method{} retains neural networks' universal approximation capabilities. We demonstrate its versatility and effectiveness across various applications: learning with piecewise constraints, learning optimization solvers with guaranteed feasibility, and optimizing control policies in safety-critical systems.\footnote{The code is available at \url{https://github.com/azizanlab/hardnet}.}
\end{abstract}


\section{Introduction}

Neural networks are widely adopted for their generalization capabilities and their ability to model highly nonlinear functions in high-dimensional spaces. With their increasing proliferation, it has become more important to be able to impose constraints on neural networks in many applications. By incorporating domain knowledge about input-output relationships through constraints, we can enhance generalization, particularly when available data is limited \citep{pathak2015constrained, oktay2017anatomically, raissi2019physics}. These constraints introduce inductive biases that guide the model's learning process toward plausible solutions that adhere to known properties of the problem domain, potentially reducing overfitting. Consequently, neural networks can more effectively capture underlying patterns and make accurate predictions on unseen data, despite scarce training samples.

Moreover, adherence to specific requirements is critical in many practical applications. For instance, in robotics, this could translate to imposing collision avoidance or ensuring configurations remain within a valid motion manifold \citep{ding2014multilayer, wang2023human, ryu2022equivariant, huang2022equivariant}. In geometric learning, this could mean imposing a manifold constraint \citep{lin2008riemannian, simeonov2022neural}. In financial risk management, violating constraints on the solvency of the portfolio can lead to large fines \citep{mcneil2015quantitative}. Enforcing neural network outputs to satisfy these non-negotiable rules (i.e., hard constraints) makes models more reliable, interpretable, and aligned with the underlying problem structure. 

However, introducing hard constraints can potentially limit a neural network's expressive power. To illustrate this point, consider a constraint that requires the model's output to be less than 1. One could simply restrict the model to always output a constant value less than 1, which ensures the constraint satisfaction but obviously limits the model capacity drastically. This raises the question:
\vspace{-0.2cm}
\begin{center}
\emph{Can we enforce hard constraints on neural networks without losing their expressive power?}
\end{center}\vspace{-0.2cm}
The model capacity of neural networks is often explained through the universal approximation theorem, which shows that a neural network can approximate any continuous function given a sufficiently wide/deep architecture. Demonstrating that this theorem still holds under hard constraints is essential to understanding the trade-off between constraint satisfaction and model capacity.


\paragraph{Contributions} 
We tackle the problem of enforcing hard constraints on neural networks, namely:
\begin{itemize}[leftmargin=.4cm]
\setlength\itemsep{-.07cm}
\item \textbf{We present a practical framework} called \method{} (short for hard-constrained neural network) for constructing neural networks that satisfy input-dependent constraints by construction. \method{} is, to the best of our knowledge, the first method that enables efficient and differentiable enforcement of more than one input-dependent inequality constraint. It allows for unconstrained optimization of the networks' parameters with standard algorithms.
\item \textbf{We prove universal approximation theorems} for our method, showing that despite enforcing the hard constraints, our construction retains the expressive power of neural networks, i.e., it provably does not overconstrain the model.
\item \textbf{We demonstrate the utility} of our method on a variety of scenarios where it is critical to satisfy hard constraints---learning with piecewise constraints, learning optimization solvers with guaranteed feasibility, and optimizing control policies in safety-critical systems.
\item \textbf{We provide the first systematic taxonomy} and comparative analysis of hard-constrained neural networks (Table~\ref{tab:comp})---aligning constraint type, input-dependence, guarantees, cost, and expressivity.
\end{itemize}

\section{Related Work}

\paragraph{Neural Networks with Soft Constraints}
Early approaches used data augmentation or domain randomization to structure the dataset to satisfy the necessary constraints before training the neural network. Other initial directions focused on introducing the constraints as \textit{soft} penalties \citep{marquez2017imposing, dener2020training} into the cost function of the neural network along with penalty weights or Lagrange multipliers as hyperparameters. \citet{raissi2019physics, LI202460} leveraged this idea in their work on physics-informed neural networks (PINNs) to enforce that the output satisfies a given differential equation.
In parallel to these penalty-based and augmented Lagrangian methods, \citet{chamon2020probably} \cite{chamon2023constrained} proposed rigorous methods that optimize over both primal and dual variables with distributional guarantees based on the PAC-learning framework. \citet{hounie2023resilient} extended this approach by adaptively relaxing constraints to find a better compromise between the objective and the constraints.
While soft-constraint methods are useful for incentivizing the desirable behavior in the model, their main limitation is that they do not ensure constraint satisfaction for arbitrary inputs, especially those far from the training distribution. 

\begin{table*}
    \caption{Comparison of methods enforcing hard constraints on neural networks for the target function $y=f(x)\in\mathbb{R}^{n_\textup{out}}$. 
    \methodaff{} is the only method to enforce input-dependent constraints provably and efficiently with universal approximation guarantees. (Eq.: Equality, Ineq.: Inequality)}
    
    \label{tab:comp}
    \centering
    \renewcommand{\arraystretch}{1.09}
    \newcommand{\colorpos}{\cellcolor[RGB]{230,255,230}}
    \newcommand{\colorsemipos}{\cellcolor[RGB]{245,255,255}}
    \newcommand{\colorneg}{\cellcolor[RGB]{255,230,230}}
    \setlength{\tabcolsep}{2pt}
    \small
    \resizebox{\linewidth}{!}{%
    \begin{tabular}{c|c c c c c c c}
        \toprule
        Method & Constraint &
        \begin{tabular}{@{}c@{}}
        Input\\Depend.
        \end{tabular} & 
        \begin{tabular}{@{}c@{}}
        Support\\Eq./Ineq.
        \end{tabular}&
        \begin{tabular}{@{}c@{}}
        Satisfaction\\Guarantee
        \end{tabular} & Computation& 
        \begin{tabular}{@{}c@{}}
        Universal\\Approx.
        \end{tabular}\\
        \midrule
        \arrayrulecolor{black!20}

        Soft-Constrained & Any & \colorpos Yes & \colorpos Both & \colorneg No & 
        \colorpos Closed-Form & \colorpos Yes \\
        \hdashline
        \citet{frerix2020homogeneous} & $Ay\leq0$ &
        \colorneg No & \colorpos Both &
        \colorpos Always & \colorpos Closed-Form & \colorneg Unknown \\
        \hdashline
        \begin{tabular}{@{}c@{}}
        LinSATNet\\
        \citep{wang2023linsatnet}
        \end{tabular}
        & \begin{tabular}{@{}c@{}}
        $A_1 y\leq b_1, A_2 y\geq b_2$\\
        $(y\in[0,1]^{n_\textup{out}}$, $A_*,b_*\geq0)$
        \end{tabular} & \colorneg No & \colorpos Both & \colorsemipos Asymptotic & 
        \colorneg Iterative & \colorneg Unknown \\
        \hdashline
        C-DGM~\citep{stoian2024how}
        & $Ay\leq b$ & \colorneg No & \colorpos Both & \colorpos Always & 
        \colorpos Closed-Form & \colorneg Unknown \\
        \hdashline
        \begin{tabular}{@{}c@{}}
        RAYEN\\
        \citep{tordesillas2023rayen}
        \end{tabular} & 
        \begin{tabular}{@{}c@{}}
        $y\in\mathcal{C}$ ($\mathcal{C}$: linear,\\ quadratic, SOC, LMI)
        \end{tabular} & \colorneg No & \colorpos Both & \colorpos Always & 
        \colorpos Closed-Form & \colorneg Unknown\\
        \hdashline
        \begin{tabular}{@{}c@{}}
        POLICE\\
        \citep{balestriero2023police}
        \end{tabular} &
        $y=Ax+b \;\; \forall x\in R$
        & \colorpos Yes & \colorneg Eq. Only & \colorpos Always & \colorpos Closed-Form & \colorneg Unknown\\
        \hdashline
        \begin{tabular}{@{}c@{}}
        KKT-hPINN\\
        \citep{chen2024physics}
        \end{tabular} &
        \begin{tabular}{@{}c@{}}
        $Ax+By=b$\\(\# constraints $\leq n_\textup{out}$)
        \end{tabular}
        & \colorpos Yes & \colorneg Eq. Only & \colorpos Always & \colorpos Closed-Form & \colorneg Unknown \\
        \hdashline
        ACnet~\citep{beucler2021enforcing} &
        \begin{tabular}{@{}c@{}}
        $h_x(y)=0$\\
        (\# constraints $\leq n_\textup{out}$)
        \end{tabular}
        & \colorpos Yes & \colorneg Eq. Only & \colorpos Always & \colorpos Closed-Form & \colorneg Unknown \\
        \hdashline
        DC3~\citep{donti2021dc3} &
        $g_x(y)\leq0, h_x(y)=0$ & \colorpos Yes & \colorpos Both & \colorsemipos
        \hspace{-.1cm}\begin{tabular}{@{}c@{}}
        Asymptotic\\
        for linear $g_x,h_x$
        \end{tabular}\hspace{-.1cm} & 
        \colorneg Iterative & \colorneg Unknown \\
        \hdashline
        \begin{tabular}{@{}c@{}}
        \methodaff{}\\(Ours)
        \end{tabular}
        & 
        \begin{tabular}{@{}c@{}}
        $b^l(x)\leq A(x)y\leq b^u(x)$\\
        (\# constraints $\leq 2n_\textup{out}$)
        \end{tabular}
        & \colorpos Yes & \colorpos Both & \colorpos Always & \colorpos Closed-Form & \colorpos Yes\\
        \arrayrulecolor{black}
        \bottomrule
    \end{tabular}
    }
\end{table*}

\paragraph{Neural Networks with Hard Constraints}
Some conventional neural network components can already enforce specific types of hard constraints. For instance, sigmoids can impose lower and upper bounds, softmax layers enforce simplex constraints, and ReLU layers are projections onto the positive orthant. The convolution layer in ConvNets encodes a translational equivariance constraint, which led to significant improvements in empirical performance. Learning new equivariances and inductive biases that accelerate learning for specific tasks is an active research area. 

Recent work has explored new architectures to (asymptotically) impose various hard constraints by either finding certain parameterizations of feasible sets or incorporating differentiable projections into neural networks, as summarized in Table~\ref{tab:comp}. 
\citet{frerix2020homogeneous} addressed homogeneous linear inequality constraints by embedding a parameterization of the feasible set in a neural network layer.
\citet{huang2021differentiable} and LinSATNet~\citep{wang2023linsatnet} introduced differentiable projection methods that iteratively refine outputs to satisfy linear constraints.
However, these iterative approaches do not guarantee constraint satisfaction within a fixed number of iterations, limiting their reliability in practice.
C-DGM~\citep{stoian2024how} enforces linear inequality constraints in generative models for tabular data
by incrementally adjusting each output component in a finite number of iterations. However, its application to input-dependent constraints is limited as it cannot efficiently handle batched data. When constraints are input-dependent, the method requires recomputing the reduced constraint sets for each input, making it computationally prohibitive. In the context of optimal power flow, \citet{chen2023end} enforces feasibility for learning optimization proxies through closed-form differentiable repair layers. While effective, this approach is restricted to specific affine constraints.

Beyond affine constraints, RAYEN \citep{tordesillas2023rayen} and \citet{konstantinov2023new} enforce certain convex constraints by parameterizing the feasible set such that the neural network output represents a translation from an interior point of the convex feasible region. However, these methods are limited to constraints dependent only on the output, and not the input. Extending these methods to input-dependent constraints is challenging because it requires finding different parameterizations for each input, such as determining a new interior point for every feasible set.

Another line of work considers hard constraints that depend on both input and output.
POLICE~\citep{balestriero2023police} enforces the output to be an affine function of the input in specific regions by reformulating the neural networks as continuous piecewise affine mappings. 
KKT-hPINN~\citep{chen2024physics} handles more general affine equality constraints by projecting the output to the feasible set where the projection is computed using KKT conditions. ACnet~\citep{beucler2021enforcing} enforces nonlinear equality constraints by transforming them into affine constraints for redefined inputs and outputs.
However, these methods are restricted to equality constraints.
DC3~\citep{donti2021dc3} tackles more general nonlinear constraints by reducing inequality constraints violations via gradient-based methods over the manifold where equality constraints are satisfied. However, it does not guarantee constraint satisfaction in general and is sensitive to the number of gradient steps and the step size, which require fine-tuning.

More closely related to our framework, methods to enforce a single affine inequality constraint have been proposed in the control literature:
\citet{kolter2019learning} presented a framework for learning a stable dynamical model that satisfies a Lyapunov stability constraint. Based on this method, \citet{min2023data} presented the CoILS framework to learn a stabilizing control policy for an unknown control system by enforcing a control Lyapunov stability constraint. Our work generalizes the ideas used in these works to impose more general affine/convex constraints while proving universal approximation guarantees that are absent in prior works;
On the theoretical front, \citet{kratsios2021universal} presented a constrained universal approximation theorem for \textit{probabilistic} transformers whose outputs are constrained to be in a feasible set. However, their contribution is primarily theoretical, and they do not present a method for learning such a probabilistic transformer. 

\paragraph{Formal Verification of Neural Networks} Verifying whether a provided neural network (after training) always satisfies a set of constraints for a certain set of inputs is a well-studied subject. \citet{albarghouthi2021introduction} provide a comprehensive summary of the constraint-based and abstraction-based approaches to verification. Constraint-based verifiers are often both sound and complete but they have not scaled to practical neural networks, whereas abstraction-based techniques are approximate verifiers which are sound but often incomplete \citep{bunel2018a, elboher2020an, brown2022unified, tjeng2018evaluating, OPT-035, fazlyab2020safety, qin2018verification, ehlers2017formal}. Other approaches have focused on formally verified exploration and policy learning for reinforcement learning \citep{bastani2018verifiable, anderson2020neurosymbolic, wabersich2022probabilistic}. Contrary to most formal verification methods, which take a pre-trained network and verify that its output always satisfies the desired constraints, our method guarantees constraint satisfaction \textit{by construction} throughout the training.

To address constraint violations in trained models after verification, some safe reinforcement learning methods use shielding mechanisms by overriding unsafe decisions using a backup policy based on reachability analysis~\citep{shao2021reachability, bastani2021safe}. While effective, this setup introduces a hard separation between learning and enforcement, often sacrificing performance and limiting joint optimization. Shielding is typically not differentiable and cannot be integrated into training. Similarly, editing methods modify trained networks post hoc to enforce output constraints by solving relaxed optimization problems over the model parameters~\citep{sotoudeh2021provable, tao2024provable}. Though recent works demonstrate their application during training, these techniques generally target input-independent constraints and are architecture-dependent.

\paragraph{Neuro-Symbolic AI} \method{} also aligns with the objectives of Neuro-symbolic AI, a field that has gained significant attention in recent years for its ability to integrate complex background knowledge into deep learning models. Unlike \method{}, which focuses on algebraic constraints, the neuro-symbolic AI literature primarily addresses logical constraints.
A common approach in this field is to \textit{softly} impose constraints during training by introducing penalty terms into the loss function to discourage constraint violations~\citep{xu2018semantic, fischer2019dl2, badreddine2022logic, stoian2023exploiting}. While these methods are straightforward to implement, they do not guarantee constraint satisfaction.
In contrast, works such as \citet{giunchiglia2020coherent, ahmed2022semantic, giunchiglia2024ccn+} ensure constraints are satisfied by embedding them into the predictive layer, thus guaranteeing compliance by construction. Another line of research maps neural network outputs into logical predicates, ensuring constraint satisfaction through reasoning on these predicates~\citep{manhaeve2018deepproblog, pryor2023neupsl, van2023nesi}.

\section{Preliminaries}
\label{section:preliminaries}
\subsection{Notation}
For $p\in[1,\infty)$, $\|v\|_p$ denotes the $\ell^p$-norm for a vector $v\in\mathbb{R}^m$, and $\|A\|_p$ denotes the operator norm for a matrix $A\in\mathbb{R}^{k\times m}$ induced by the $\ell^p$-norm, \emph{i.e.}, $\|A\|_p=\sup_{w\neq0} \|Aw\|_p/\|w\|_p$.
$v_{(i)}\in\mathbb{R}$ denotes the $i$-th component of $v$.
$[A; B]$ denotes the row-wise concatenation of the matrices $A$ and $B$.

For a domain $\mathcal{X}\subset\mathbb{R}^{n_\textup{in}}$ and a codomain $\mathcal{Y}\subset\mathbb{R}^{n_\textup{out}}$, let $\mathcal{C}(\mathcal{X},\mathcal{Y})$ be the class of continuous functions from $\mathcal{X}$ to $\mathcal{Y}$ endowed with the sup-norm: $\|f\|_\infty:=\sup_{x\in\mathcal{X}} \|f(x)\|_\infty$. Similarly, $L^p(\mathcal{X},\mathcal{Y})$ denotes the class of $L^p$ functions from $\mathcal{X}$ to $\mathcal{Y}$ with the $L^p$-norm: $\|f\|_p:=(\int_\mathcal{X} \|f(x)\|_p^p dx)^{\frac{1}{p}}$. 
For function classes $\mathcal{F}_1, \mathcal{F}_2 \subset \mathcal{C}(\mathcal{X},\mathcal{Y})$ (respectively, $\mathcal{F}_1, \mathcal{F}_2 \subset L^p(\mathcal{X},\mathcal{Y})$), we say $\mathcal{F}_1$ \textit{universally approximates} (or \textit{is dense in}) $\mathcal{F}_2$ if for any $f_2\in\mathcal{F}_2$ and $\epsilon>0$, there exists $f_1\in\mathcal{F}_1$ such that $\|f_2-f_1\|_\infty\leq\epsilon$ (respectively, $\|f_2-f_1\|_p\leq\epsilon$).
For a neural network, its depth and width are defined as the total number of layers and the maximum number of neurons in any single layer, respectively.

\subsection{Universal Approximation Theorem}
The universal approximation property is a foundational concept in understanding the capabilities of neural networks in various applications. Classical results reveal that shallow neural networks with arbitrary width can approximate any continuous function defined on a compact set as formalized in the following theorem~\citep{cybenko1989approximation, hornik1989multilayer, leshno1993multilayer, pinkus1999approximation}:
\begin{theorem}[Universal Approximation Theorem for Shallow Networks]
\label{thm:uat_shallow}
    Let $\rho\in\linebreak\mathcal{C}(\mathbb{R},\mathbb{R})$ and $\mathcal{K}\in\mathbb{R}$ be a compact set. Then, depth-two neural networks with $\rho$ activation function \textit{universally approximate} $\mathcal{C}(\mathcal{K},\mathbb{R})$ if and only if $\rho$ is nonpolynomial.
\end{theorem}

To further understand the success of deep learning, the universal approximation property for deep and narrow neural networks has also been studied in the literature~\citep{lu2017expressive, hanin2017approximating, kidger2020universal, park2021minimum}. Interesting results show that a critical threshold exists on the width of deep networks that attain the universal approximation property. For instance, deep networks with ReLU activation function with a certain minimum width can approximate any $L^p$ function as described in the following theorem~\citep[Thm. 1]{park2021minimum}:
\begin{theorem}[Universal Approximation Theorem for Deep Networks]
\label{thm:uat_deep}
    For any $p\in[1,\infty)$, $w$-width neural networks with \textup{$\relu{}$} activation function \textit{universally approximate} $L^p(\mathbb{R}^{n_\textup{in}},\mathbb{R}^{n_\textup{out}})$ if and only if $w\geq\max\{n_\textup{in}+1,n_\textup{out}\}$.
\end{theorem}

Despite these powerful approximation guarantees, they fall short when neural networks are required to satisfy hard constraints, such as physical laws or safety requirements. These theorems ensure that a neural network can approximate a target function arbitrarily closely but do not guarantee adherence to necessary constraints. Consequently, even if the target function satisfies specific hard constraints, the neural network approximator might violate them---especially in regions where the target function barely meets the constraints.
This shortcoming is particularly problematic for applications that demand strict compliance with non-negotiable domain-specific rules.
Therefore, ensuring that neural networks can both approximate target functions accurately and rigorously satisfy hard constraints remains a critical challenge for their deployment in practical applications.

\begin{figure}[t]
    \centering
    \includegraphics[width=.75\linewidth]{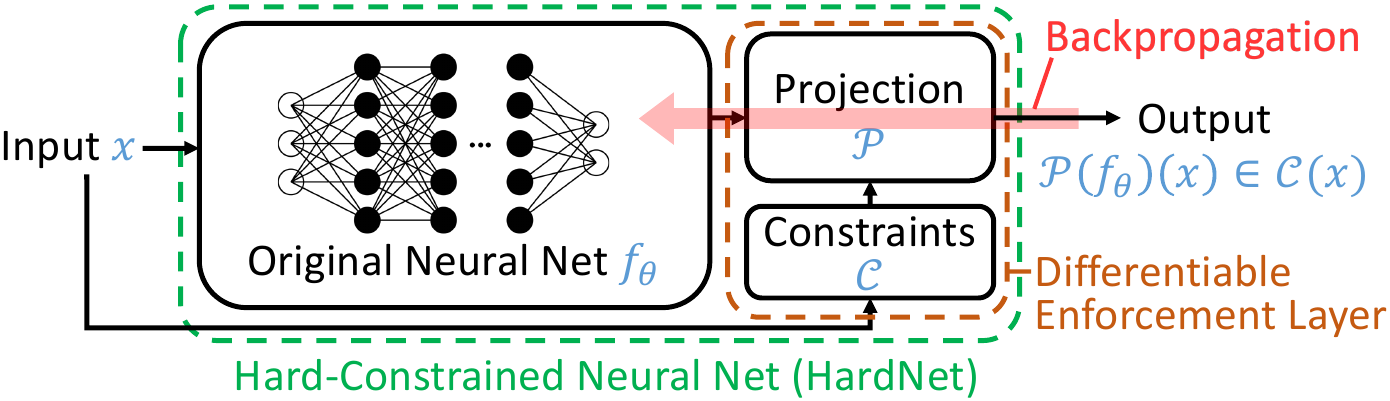}
    \caption{Schematic of \method{}. Its differentiable enforcement layer allows unconstrained end-to-end optimization of the network parameters using standard algorithms while guaranteeing satisfaction with input-dependent constraints by construction. The layer can be applied to any neural networks.}
    \label{fig:schematic}
\end{figure}

\section{\method{}: Hard-Constrained Neural Network}
\label{section:hardnet}
In this section, we present a practical framework, \method{}, shown in Figure~\ref{fig:schematic}, for enforcing input-dependent hard constraints on neural networks while retaining their universal approximation properties. In a nutshell, for a parameterized (neural network) function $f_\theta:\mathcal{X}\subset\mathbb{R}^{n_\textup{in}}\rightarrow\mathbb{R}^{n_\textup{out}}$, we ensure the satisfaction of given constraints by appending a differentiable enforcement layer with a projection $\mathcal{P}$ to $f_\theta$. This results in the projected function $\mathcal{P}(f_\theta):\mathcal{X}\rightarrow\mathbb{R}^{n_\textup{out}}$ meeting the required constraints while allowing its output to be backpropagated through to train the model via gradient-based algorithms.
Importantly, we show that the proposed architecture has universal approximation guarantees, i.e., it universally approximates the class of functions that satisfy the constraints. 

A key challenge in this approach is devising a differentiable projection that has efficient forward and backward passes. For instance, consider affine constraints $A(x)f(x)\leq b(x) \;\forall x\in\mathcal{X}$ for a function $f:\mathcal{X}\rightarrow\mathbb{R}^{n_\textup{out}}$ with $A(x)\in\mathbb{R}^{n_\textup{c}\times n_\textup{out}}$ and $b(x)\in\mathbb{R}^{n_\textup{c}}$, one would have
\begin{equation}\label{eq:projection_ineq}
    \mathcal{P}(f_\theta)(x)
    = \argmin_{z\in\mathbb{R}^{n_\textup{out}}} \|z\!-\! f_\theta(x)\|_2 \textup{ s.t. } A(x) z \leq b(x).
\end{equation}
Although the constraints are affine, this optimization \emph{does not admit a closed-form solution} in general for more than one constraint, making it computationally expensive---especially as it needs to be computed for every sample $x$ (during training/inference) and every parameter $\theta$ (during training).

In case of a single constraint $a(x)^\top f(x)\leq b(x) \;\;\forall x\in\mathcal{X}$
with $a(x)\in\mathbb{R}^{n_\textup{out}}$ and $b(x)\in\mathbb{R}$, recent work in the control literature---for instance, \citet{kolter2019learning} for learning stable dynamics and \citet{donti2021enforcing, min2023data} for learning stabilizing controllers---has adopted the following closed-form solution, which is differentiable almost everywhere:
\begin{align}
    \mathcal{P}(f_\theta)(x)
    &= \argmin_{z\in\mathbb{R}^{n_\textup{out}}} 
    \|z- f_\theta(x)\|_2 \textup{ s.t. } a(x)^\top z \leq b(x)
    \label{eq:proj_affine_single_opt}\\
    &= f_\theta(x)- \dfrac{a(x)}{\|a(x)\|^2}\relu{}\big(a(x)^\top f_\theta(x)-b(x)\big).
    \label{eq:proj_affine_single_closed}
\end{align}
\begin{example}
$a(x)\!=\![-1; x]$ and $b(x)\!=\!0$ encode the constraint $(f(x))_{(0)}\!\geq\! x (f(x))_{(1)}$ on $f\!:\!\mathbb{R}\!\rightarrow\!\mathbb{R}^2$. Then, a sample $f_\theta(1)\!=\![3;5]$ is projected to $\mathcal{P}(f_\theta)(1)\!=\![4;4]$, satisfying the constraint.
\end{example}

Nonetheless, the formulation is limited to enforcing \emph{only a single} inequality constraint. Moreover, its empirical success in learning the desired functions has not been theoretically understood. To that end, we propose \methodaff{}, \emph{the first method, to the best of our knowledge, that enables efficient (closed-form) and differentiable enforcement of more than one input-dependent inequality constraint.}
In addition, we provide a rigorous explanation for its expressivity through universal approximation guarantees.
Then, we discuss \methodcvx{} as a conceptual framework to satisfy general input-dependent convex constraints exploiting differentiable convex optimization solvers.

\subsection{\methodaff{}: Imposing Input-Dependent Affine Constraints}
\label{section:method_affine}

Suppose we have multiple input-dependent affine constraints in an aggregated form:
\begin{equation}\label{eq:constraint_affine_multi}
    b^l(x)\leq A(x)f(x)\leq b^u(x)\;\;\;\forall x\in\mathcal{X},
\end{equation}
where $A(x)\in\mathbb{R}^{n_\textup{c}\times n_\textup{out}}$ and $b^l(x),b^u(x)\in\mathbb{R}^{n_\textup{c}}$ for $2n_\textup{c}$ inequality constraints.
\begin{remark}
    The constraint form in~\eqref{eq:constraint_affine_multi} is general and includes equality constraints by setting $b^l=b^u$.
    Another approach to enforcing equality constraints is to have the neural network predict only a subset of the outputs, with the remaining components computed to satisfy the equality constraints, as in \citet{beucler2021enforcing} and \citet{donti2021dc3}. However, that method may fail if the chosen subset is invalid for certain inputs.
    For example, the constraint $[x\;x\!-\!1]f(x)\!=\!1$ on $f\!:\mathbb{R}\!\rightarrow\!\mathbb{R}^2$ prevents either component from being a free variable at both $x\!=\!0\text{ and }1$. See Appendix~\ref{appendix:equality} for incorporating this approach into our method.
\end{remark}
\begin{remark}
    One-sided inequality constraints can be represented by setting $b^l\!=\!-\infty$ or $b^u\!=\!\infty$. In addition, different types of constraints (inequality and equality) can be combined by specifying $b^l$ and $b^u$ component-wise.
\end{remark}
\begin{assumption}\label{asmp:affine}
    For each $x\in\mathcal{X}$,
    \romannumeral 1) there exists $y\in\mathbb{R}^{n_\textup{out}}$ that satisfies the constraints in~\eqref{eq:constraint_affine_multi}, and
    \;\romannumeral 2) $A(x)$ has full row rank.
\end{assumption}
The first assumption says that the constraints~\eqref{eq:constraint_affine_multi} are feasible, while the second one requires $n_\textup{c}\leq n_\textup{out}$.
Under the assumptions, we propose \methodaff{} by developing a novel closed-form projection that enforces the constraints~\eqref{eq:constraint_affine_multi} as below:
\begin{equation}\label{eq:hardnet_aff}
    \methodaff{}\!:
    \mathcal{P}(f_\theta)(x)
    \!=\!
    f_\theta(x) + A(x)^+ \big[\relu{}\big(b^l(x)-A(x)f_\theta(x)\big) - \relu{}\big(A(x)f_\theta(x)-b^u(x)\big)\big],
\end{equation}
for all $x\in\mathcal{X}$
where $M^+:=M^\top (MM^\top)^{-1}$ denotes the pseudoinverse of a matrix $M$.
Note that the projection in \eqref{eq:hardnet_aff} generalizes the single-constraint case in \eqref{eq:proj_affine_single_closed} with $n_c=1$ and $b^l=-\infty$. However, unlike \eqref{eq:proj_affine_single_closed}, \methodaff{} in general does not perform the minimum $\ell^2$-norm projection as in~\eqref{eq:proj_affine_single_opt}. Instead, we can characterize its projection through the following optimization problem; see Appendix~\ref{appendix:affine_opt} for a proof.
\begin{restatable}{proposition}{AffOpt}\label{prop:affine_opt}
    Under Assumption~\ref{asmp:affine}, for any parameterized function $f_\theta:\mathcal{X}\rightarrow\mathbb{R}^{n_\textup{out}}$, \methodaff{} in~\eqref{eq:hardnet_aff} satisfies
    \begin{equation}\label{eq:affine_opt}
    \begin{aligned}
        \mathcal{P}(f_\theta)(x)
    = \argmin_{z\in\mathbb{R}^{n_\textup{out}}} & \|z- f_\theta(x)\|_2 \\
    \textup{ s.t.} &\;
    z\in\Big\{\argmin_{y\in\mathbb{R}^{n_\textup{out}}}
    \big\|A(x)\big(y- f_\theta(x)\big)\big\|_2 
    \textup{ s.t. } b^l(x) \leq A(x) y \leq b^u(x)\Big\}.
    \end{aligned}
    \end{equation}
\end{restatable}
\noindent 
Thus, \methodaff{} satisfies the constraint~\eqref{eq:constraint_affine_multi} while minimally altering the output from the plain output $f_\theta(x)$ in the sense of \eqref{eq:affine_opt}.
We note that the inner optimization in~\eqref{eq:affine_opt} has infinitely many solutions when $A(x)$ has a non-zero null space (i.e., $n_c<n_\textup{out}$). In such cases, \methodaff{} chooses the solution that is closest (in Euclidean distance) to the plain output $f_\theta(x)$. On the other hand, for a square matrix $A(x)$, the inner optimization has a unique solution while the square of its objective function is the Bregman divergence generated from the function $\psi_x(y)=\| A(x) y\|_2^2$.

\begin{figure}[t]
    \centering
    \includegraphics[width=.75\linewidth]{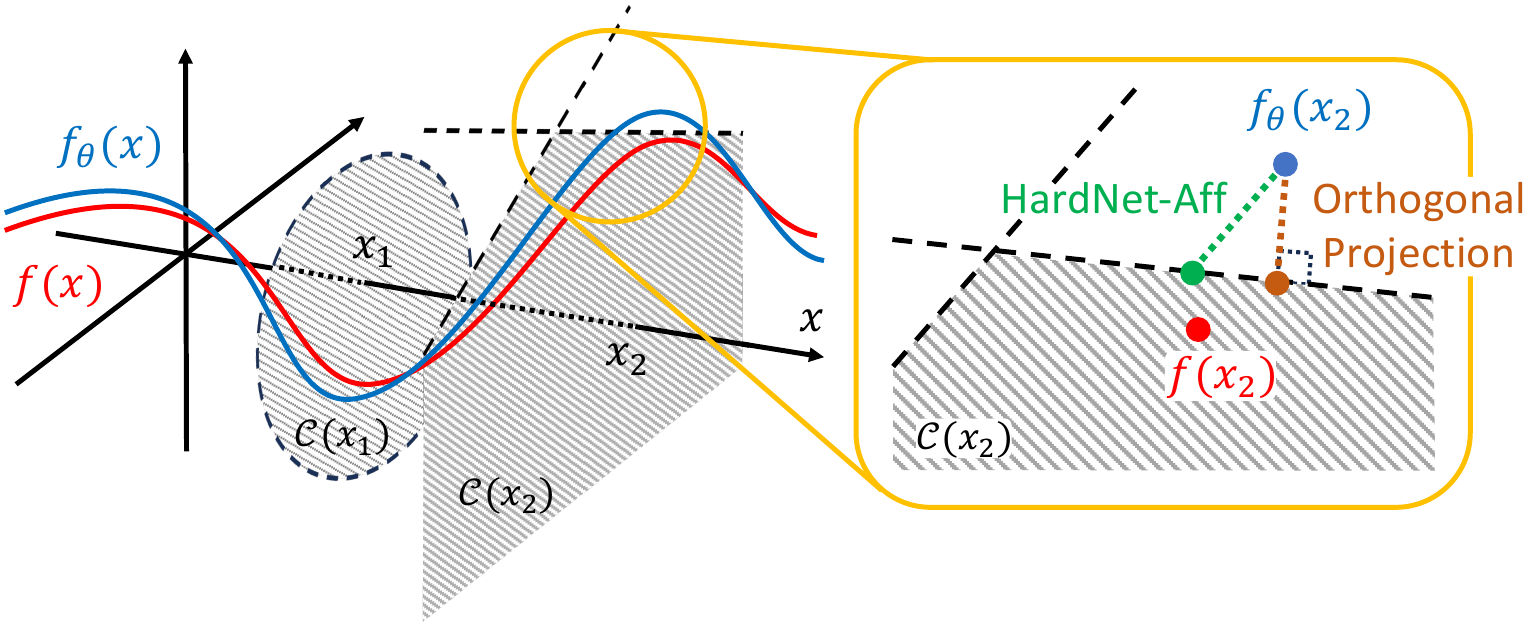}
    \caption{Illustration of input-dependent constraints and projections performed by \method{}. A target function $f\!:\!\mathbb{R}\!\rightarrow\!\mathbb{R}^2$ satisfies hard constraints $f(x)\!\in\!\mathcal{C}(x)$ for each $x\!\in\!\mathbb{R}$. The feasible set $\mathcal{C}(x)$ is visualized as the gray area for two sample inputs $x_1$ and $x_2$. While the function $f_\theta$ closely approximates $f$, it violates the constraints. \methodaff{} projects the violated output onto the feasible set in parallel to the boundaries of the satisfied constraints.
    In contrast, the minimum $\ell^2$-norm optimization in~\eqref{eq:projection_ineq} projects the output orthogonally to the closest boundary.}
    \label{fig:projection}
    \vspace{-0.2cm}
\end{figure}

In addition to the optimization formulation, \methodaff{} can also be understood geometrically through the following property; see Appendix~\ref{appendix:affine_prop} for a proof.
\begin{restatable}[Parallel Projection]{proposition}{AffProp}\label{prop:affine_prop}
    Under Assumption~\ref{asmp:affine}, for any parameterized function $f_\theta:\mathcal{X}\rightarrow\mathbb{R}^{n_\textup{out}}$, for each $i$-th row $a_i\in\mathbb{R}^{n_\textup{c}}$ of $A(x)$, \methodaff{} in~\eqref{eq:hardnet_aff} satisfies
    \begin{equation}
    a_i^\top \mathcal{P}(f_\theta)(x) = \begin{cases}
        b^l_{(i)}(x) & \hspace{-.5cm}\textup{if }\; a_i^\top f_\theta(x) < b^l_{(i)}(x)\\
        b^u_{(i)}(x) &\hspace{-.5cm}\textup{if }\; a_i^\top f_\theta(x) > b^u_{(i)}(x)\\
        a_i^\top f_\theta(x) & \textup{o.w.}
    \end{cases}
    \;\;\;\text{ for all }\; x\in\mathcal{X}.
    \end{equation}
\end{restatable}
\noindent If $f_\theta(x)$ violates any constraints, it is projected onto the boundary of the feasible set ($a_i^\top \mathcal{P}(f_\theta)(x)=b^l_{(i)}(x)\textup{ or }b^u_{(i)}(x)$). Notably, the projection alters the output in a direction parallel to the boundary of each satisfied constraint's feasible set ($a_i^\top \mathcal{P}(f_\theta)(x)=a_i^\top f_\theta(x)$). This parallel projection is illustrated in Fig.~\ref{fig:projection}.

While \methodaff{} guarantees the satisfaction of the hard constraints~\eqref{eq:constraint_affine_multi}, it should not lose the neural networks' expressivity for practical deployment. To that end, we rigorously show that \methodaff{} preserves the neural networks' expressivity by the following theorem; see Appendix~\ref{appendix:affine_uat} for a proof.
\begin{restatable}{theorem}{AffUAT}\label{thm:affine_uat}
    Consider input-dependent constraints~\eqref{eq:constraint_affine_multi} that satisfy assumption~\ref{asmp:affine}. Suppose $\mathcal{X}\subset\mathbb{R}^{n_\textup{in}}$ is compact, and $A(x)$ is continuous over $\mathcal{X}$.
    Then, for any function classes $\mathcal{F}_\textup{NN}, \mathcal{F} \subset \mathcal{C}(\mathcal{X},\mathbb{R}^{n_\textup{out}})$ (or $\mathcal{F}_\textup{NN}, \mathcal{F} \subset L^p(\mathcal{X},\mathbb{R}^{n_\textup{out}})$ for any $p\in[1,\infty)$) and the projection $\mathcal{P}$ of \methodaff{} in~\eqref{eq:hardnet_aff},
    if $\mathcal{F}_\textup{NN}$ universally approximates $\mathcal{F}$,
    $\mathcal{F}_\textup{\methodaff{}}:=\{\mathcal{P}(f_\textup{NN})| f_\textup{NN}\in\mathcal{F}_\textup{NN}\}$ universally approximates $\mathcal{F}_\textup{target}:=\{f_t\in\mathcal{F}| f_t \text{ satisfies~\eqref{eq:constraint_affine_multi}}\}$.
\end{restatable}
Considering $\mathcal{F}_\textup{NN}$ as the function class of plain neural networks, this theorem shows that \methodaff{} preserves their expressivity over $\mathcal{F}$ under the constraints~\eqref{eq:constraint_affine_multi}. The main idea behind this theorem is that for \methodaff{} in~\eqref{eq:hardnet_aff}, $\|f-\mathcal{P}(f_\theta)\|$ could be bounded in terms of $\|f-f_\theta\|$. By selecting $f_\theta$ such that $\|f-f_\theta\|$ is arbitrarily small, we can make $\mathcal{P}(f_\theta)$ approach the target function $f$ as closely as desired. The existence of such an $f_\theta$ can be guaranteed by existing universal approximation theorems on plain neural networks. For instance, if we utilize Theorem~\ref{thm:uat_deep} in Theorem~\ref{thm:affine_uat}, we can obtain the following universal approximation theorem for \methodaff{}.
\begin{corollary}\label{cor:affine_uat}
    Suppose the assumptions of Theorem~\ref{thm:affine_uat} hold.
    Then, for any $p\in[1,\infty)$, \methodaff{} with $w$-width \textup{$\relu{}$} neural networks defined in~\eqref{eq:hardnet_aff} universally approximates $\mathcal{F}_\textup{target}:=\{f_t\in L^p(\mathcal{X},\mathbb{R}^{n_\textup{out}})| f_t \text{ satisfies~\eqref{eq:constraint_affine_multi}}\}$ if $w\geq\max\{n_\textup{in}+1,n_\textup{out}\}$.
\end{corollary}

\subsection{\methodcvx{}: Imposing Input-Dependent Convex Constraints}

Going beyond affine constraints, we discuss \methodcvx{} as a conceptual framework for enforcing general input-dependent convex constraints $f(x)\in\mathcal{C}(x) \;\forall x\in\mathcal{X}$
where $\mathcal{C}(x)\subset\mathbb{R}^{n_\textup{out}}$ is a convex set. Unlike the affine case, the closed-form projection from the single-constraint case in~\eqref{eq:proj_affine_single_closed} does not extend to general convex constraints.
Thus, we present \methodcvx{} by generalizing the optimization-based projection in~\eqref{eq:proj_affine_single_opt} as below:
\begin{equation}\label{eq:hardnet_cvx}
    \methodcvx{}:
    \mathcal{P}(f_\theta)(x) = 
    \argmin_{z\in\mathbb{R}^{n_\textup{out}}} 
    \; \|z- f_\theta(x)\|_2 
    \;\textup{ s.t. } z\in\mathcal{C}(x),
\end{equation}
for all $x\in\mathcal{X}$. While no general closed-form solution for this optimization problem exists, we can employ differentiable convex optimization solvers for an implementation of \methodcvx{} such as \citet{amos2017optnet} for affine constraints (when \methodaff{} cannot be applied) and \citet{agrawal2019differentiable} for more general convex constraints.
This idea was briefly mentioned by~\citet{donti2021dc3} and used as a baseline (for input-independent constraints) by~\citet{tordesillas2023rayen}. 
We present \methodcvx{} as a general framework, independent of specific implementation methods, to complement \methodaff{} and provide a unified solution for various constraint types.

As in Section~\ref{section:method_affine}, we demonstrate that \methodcvx{} preserves the expressive power of neural networks by proving the following universal approximation theorem; see Appendix~\ref{appendix:convex_uat} for a proof.
\begin{restatable}{theorem}{CvxUAT}\label{thm:convex_uat}
    Consider input-dependent sets $\mathcal{C}(x)\!\subset\!\mathbb{R}^{n_\textup{out}}$ that are convex for all $x\in\mathcal{X}\subset\mathbb{R}^{n_\textup{in}}$.
    Then, for any $p\in[1,\infty)$,
    \methodcvx{} with $w$-width \textup{$\relu{}$} neural networks defined in~\eqref{eq:hardnet_cvx} universally approximates $\mathcal{F}_\textup{target}:=\{f_t\in L^p(\mathcal{X},\mathbb{R}^{n_\textup{out}})| f_t(x)\in\mathcal{C}(x) \;\forall x\in\mathcal{X}\}$ if $w\geq\max\{n_\textup{in}+1, n_\textup{out}\}$.
\end{restatable}

\section{Experiments}
\label{section:exp}

In this section, we demonstrate the versatility and effectiveness of \methodaff{} over three constrained scenarios: learning with piecewise constraints, learning optimization solvers with guaranteed feasibility, and optimizing control policies in safety-critical systems. 

As evaluation metrics, we measure the violation of constraints in addition to the application-specific performance metrics.
For a test sample $x\!\in\!\mathcal{X}$ and $n_\textup{ineq}$ inequality constraints $g_x\big(f(x)\big)\!\leq\!0\!\in\!\mathbb{R}^{n_\textup{ineq}}$, their violation is measured with the maximum ($\leq$ max) and mean ($\leq$ mean) of $\relu{}(g_x(f(x)))$ and the number of violated constraints ($\leq$ \#). Similar quantities of $|h_x(f(x))|$ are measured for $n_\textup{eq}$ equality constraints $h_x(f(x))=0\in\mathbb{R}^{n_\textup{eq}}$. Then, they are averaged over all test samples. The inference time ($T_\textup{test}$) for the test set and the training time ($T_\textup{train}$) are also compared.

We compare \method{} with the following baselines: (\romannumeral 1) \textbf{NN}: Plain neural networks, (\romannumeral 2) \textbf{Soft}: Soft-constrained neural networks, where constraint violations are penalized by adding regularization terms $\lambda \|\relu{}(g_x(f(x_s)))\|_2^2 + \lambda \|h_x(f(x_s))\|_2^2$ to the loss function for each sample $x_s\in\mathcal{X}$, (\romannumeral 3)~\textbf{DC3}~\citep{donti2021dc3}: Similarly to \methodaff{}, DC3 approximates part of the target function with a neural network. It first augments the output to satisfy equality constraints, then corrects it using gradient descent to minimize inequality violations. DC3 backpropagates through this iterative correction procedure to train the model.
All methods use 3-layer fully connected neural networks with 200 neurons per hidden layer and $\relu{}$ activation. The results are produced with Intel Xeon Gold 6248 and NVidia Volta V100.

\begin{figure}[t]
    \centering
    \includegraphics[width=0.49\linewidth]{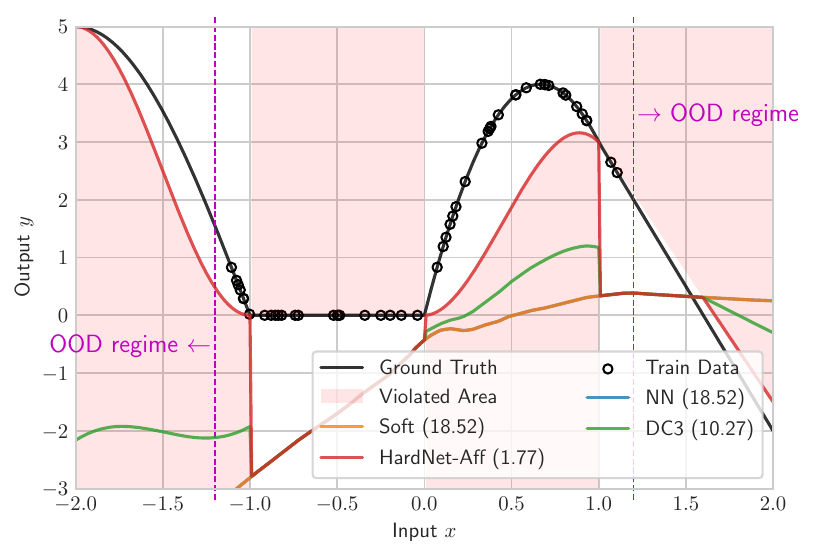}
    \includegraphics[width=0.49\linewidth]{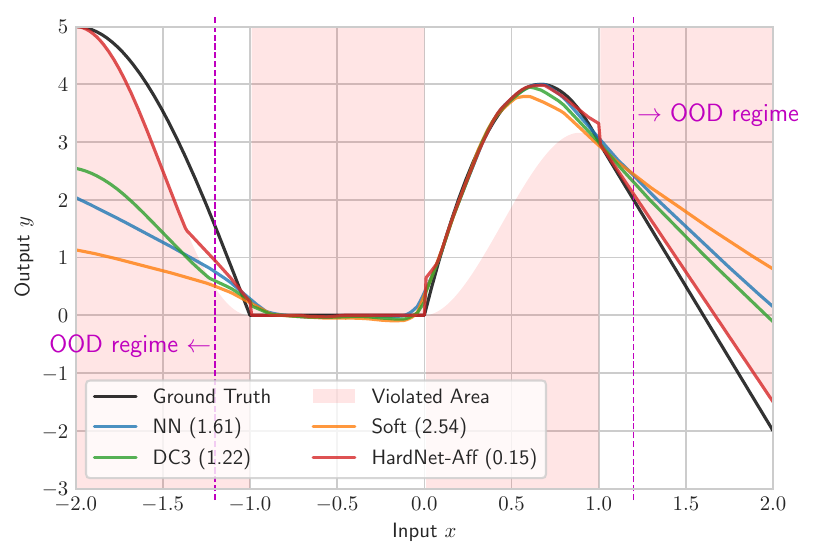}
    \caption{Learned functions at the initial (left) and final (right) epochs with the piecewise constraints. The models are trained on the samples indicated with circles, with their MSE from the true function shown in parentheses. \methodaff{} adheres to the constraints from the start of the training and generalizes better than the baselines as it enforces constraints even in the out-of-distribution (OOD) regime. On the other hand, the baselines violate the constraints throughout the training.}
    \label{fig:exp_pwc}
\end{figure}

\begin{table}[t]
    \centering
    \caption{Results for learning with the piecewise constraints. \methodaff{} generalizes better than the baselines with the smallest MSE from the true function without any constraint violation. The max and mean of constraint violations are computed over 401 test samples. Better (resp. worse) values are colored greener (resp. redder). Standard deviations over 5 runs are shown in parentheses.}
    \label{tab:pwc}
    \small
    \setlength{\tabcolsep}{3pt}
    \begin{tabular}{lccccc}
        \toprule
        {} & MSE & max violation & mean violation & $T_\textup{test}$ (ms) & $T_\textup{train}$ (s)\\
        \midrule
        NN & \cellcolor[RGB]{249,219,214} 1.85 (0.19) & 
        \cellcolor[RGB]{246,213,211} 3.16 (0.29) & \cellcolor[RGB]{254,254,240} 0.60 (0.04) & \cellcolor[RGB]{205,227,216} 0.14 (0.00) & \cellcolor[RGB]{205,227,216} 2.24 (0.09)\\
        Soft & \cellcolor[RGB]{237,204,211} 2.36 (0.22) &
        \cellcolor[RGB]{238,205,211} 3.70 (0.25) & \cellcolor[RGB]{254,253,239} 0.69 (0.03) & \cellcolor[RGB]{204,224,215} 0.15 (0.02) & \cellcolor[RGB]{222,240,223} 3.86 (0.03)\\
        DC3  & \cellcolor[RGB]{254,241,225} 1.15 (0.10) &
        \cellcolor[RGB]{253,228,218} 2.31 (0.23) & \cellcolor[RGB]{250,253,236} 0.43 (0.02) & \cellcolor[RGB]{238,205,211} 6.02 (0.07) & \cellcolor[RGB]{238,205,211} 15.67 (0.82)\\
        \methodaff{}  & \cellcolor[RGB]{207,231,218} 0.16 (0.01) & \cellcolor[RGB]{208,232,219} 
        0.00 (0.00) & \cellcolor[RGB]{208,232,219} 0.00 (0.00) & \cellcolor[RGB]{218,238,222} 0.88 (0.05) & \cellcolor[RGB]{243,250,229} 5.62 (0.06)\\
        \bottomrule
    \end{tabular}
\end{table}

\subsection{Learning with Piecewise Constraints}
\label{section:exp_pwc}
In this experiment, we demonstrate the efficacy of \methodaff{} and the expressive power of input-dependent constraints on a problem involving learning a function $f:[-2,2]\rightarrow\mathbb{R}$ with piecewise constraints shown in Fig.~\ref{fig:exp_pwc}. The function outputs are required to avoid specific regions defined over separate subsets of the domain $[-2,2]$. An arbitrary number of such piecewise constraints can be captured by a single input-dependent affine constraint. The models are trained on 50 labeled data points randomly sampled from $[-1.2,1.2]$; see Appendix~\ref{appendix:pwc} for details.
Additionally, we assess \methodaff{} with more complex constraints in which each regime is governed by both upper and lower bounds (or, in the degenerate case, an equality) in Appendix~\ref{appendix:pwcbox}.

As shown in Fig.~\ref{fig:exp_pwc} and Table~\ref{tab:pwc}, \methodaff{} consistently satisfies the hard constraints throughout training and achieves better generalization than the baselines, which violate these constraints. Especially at the boundaries $x\!=\!-1$ and $x=1$ in the initial epoch results, the jumps in DC3's output value, caused by DC3's correction process, insufficiently reduce the constraint violations. The performance of its iterative correction process heavily depends on the number of gradient descent steps and the step size. DC3 requires careful hyperparameter tuning unlike \methodaff{}.

\begin{table}[t]
    \centering
    \caption{Results for learning solvers of nonconvex optimization problems with 100 variables, 50 equality constraints, and 50 inequality constraints. \methodaff{} attain feasible solutions with the smallest suboptimality gap among the feasible methods. The max, mean, and the number of violations are computed out of the 50 constraints. Better (resp. worse) values are colored greener (resp. redder). Standard deviations over 5 runs are shown in parentheses.}
    \label{tab:nonconvex}
    \small
    \setlength{\tabcolsep}{1pt}
    \resizebox{\columnwidth}{!}{%
    \begin{tabular}{lccccc}
        \toprule
        {} & Obj. val & $\nleq$ max/mean/\#  & $\neq$ max/mean/\#& $T_\textup{test}$ (ms) & $T_\textup{train}$ (s) \\
        \midrule
        Optimizer & \cellcolor[RGB]{207,231,218} -14.28 (0.00) &  \cellcolor[RGB]{207,231,218}
        0.0/0.0/0.0 (0.0/0.0/0.0) &
        \cellcolor[RGB]{207,231,218} 0.0/0.0/0.0 (0.0/0.0/0.0) &  
        \cellcolor[RGB]{237,204,211} 1182.0 (3.49) & - \\
        
        NN & \cellcolor[RGB]{207,231,218} -27.43 (0.00) & 
        \cellcolor[RGB]{254,247,230} 12.1/1.1/12.0 (0.0/0.0/0.0) &  
        \cellcolor[RGB]{237,204,211} 15.1/6.4/50.0 (0.0/0.0/0.0) &   
        \cellcolor[RGB]{206,230,217} 0.32 (0.06)  & \cellcolor[RGB]{208,233,219} 78.10 (2.36)\\
        
        Soft & \cellcolor[RGB]{254,254,241} -13.13 (0.01) & \cellcolor[RGB]{207,231,218} 0.0/0.0/0.0 (0.0/0.0/0.0) &  
        \cellcolor[RGB]{237,204,211} 0.4/0.1/50.0 (0.0/0.0/0.0) &  
        \cellcolor[RGB]{207,230,218} 0.37 (0.07) & 
        \cellcolor[RGB]{208,233,219} 77.58 (0.97)\\
        
        DC3 &  \cellcolor[RGB]{254,234,221} -12.57 (0.04)
         &
        \cellcolor[RGB]{207,231,218} 0.0/0.0/0.0 (0.0/0.0/0.0) &  
        \cellcolor[RGB]{207,231,218} 0.0/0.0/0.0 (0.0/0.0/0.0) &  
        \cellcolor[RGB]{254,253,239} 9.61 (0.47) & 
        \cellcolor[RGB]{244,211,211} 3606.74 (2.45)
         \\
        
        \methodaff{}  &  
        \cellcolor[RGB]{228,243,224} -14.10 (0.01)
         & 
        \cellcolor[RGB]{207,231,218} 0.0/0.0/0.0 (0.0/0.0/0.0) & \cellcolor[RGB]{207,231,218} 0.0/0.0/0.0 (0.0/0.0/0.0) &  
        \cellcolor[RGB]{231,244,224} 6.69 (0.06) & 
        \cellcolor[RGB]{253,254,239} 1343.80 (1.39) \\
        \bottomrule
    \end{tabular}
    }
\end{table}

\subsection{Learning Optimization Solvers with Guaranteed Feasibility}
\label{section:exp_opt}

We consider learning optimization solvers with the following nonconvex optimization as in~\citep{donti2021dc3}:
\begin{equation*}
    f(x) = \argmin_y \;\; \frac{1}{2} y^\top Q y + p^\top \sin{y}
    \;\textup{ s.t. } Ay\leq b, \; Cy=x,
\end{equation*}
where $Q\in\mathbb{R}^{n_\textup{out}\times n_\textup{out}}\succeq 0, p\in\mathbb{R}^{n_\textup{out}}, A\in\mathbb{R}^{n_\textup{ineq}\times n_\textup{out}}, b\in\mathbb{R}^{n_\textup{ineq}}, C\in\mathbb{R}^{n_\textup{eq}\times n_\textup{out}}$ are constants and $\sin$ is the element-wise sine function. The target function $f$ outputs the solution of each optimization problem instance determined by the input $x\in[-1,1]^{n_\textup{eq}}$. 
The main benefit of learning this nonconvex optimization solver with neural networks is their faster inference time than optimizers based on iterative methods. To ensure that the learned neural networks provide feasible solutions, the constraints of the optimization problems are set as hard constraints.

In this experiment, we guarantee that the given constraints are feasible for all $x\in[-1,1]^{n_\textup{eq}}$ by computing a proper $b$ for randomly generated $A, C$ as described in~\citep{donti2021dc3}. Then the models are trained on 10000 unlabeled data points uniformly sampled from $[-1,1]^{n_\textup{eq}}$. For this unsupervised learning task, the loss function for each sample $x_s$ is set as $\frac{1}{2} f_\theta(x_s)^\top Q f_\theta(x_s) + p^\top \sin{f_\theta(x_s)}$. To reproduce similar results as in~\citep{donti2021dc3}, the models are equipped with additional batch normalization and dropout layers in this experiment.
As shown in Table~\ref{tab:nonconvex}, \methodaff{} consistently finds feasible solutions with a small suboptimality gap from the optimizer (IPOPT) with a much shorter inference time.

\subsection{Optimizing Control Policies in Safety-Critical Systems}

\begin{figure}
    \centering
    \includegraphics[width=.5\linewidth]{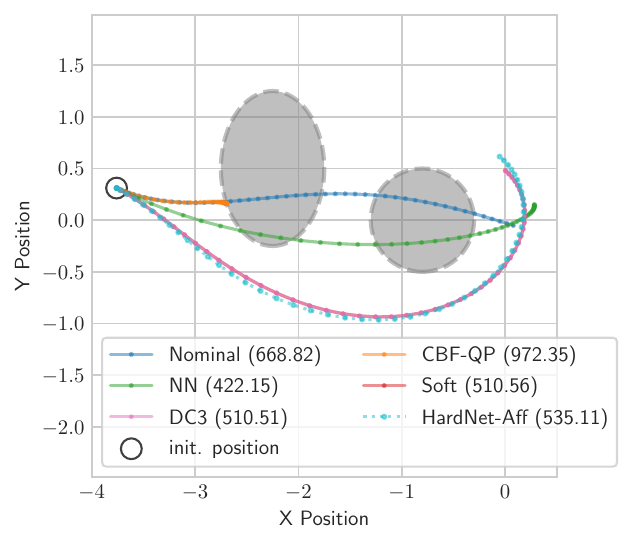}
    \caption{Simulated trajectories from a random initial state, with costs shown in parentheses. \methodaff{} avoids the obstacles while obtaining a low cost value. Even though the soft-constrained method and DC3 appear to avoid obstacles and achieve smaller costs than the other collision-free trajectories, they violate the safety constraints (which are more conservative than hitting the obstacles).
    }
    \label{fig:exp_cbf}
\end{figure}

\begin{table}
    \centering
    \caption{Results for optimizing safe control policies. \methodaff{} generates trajectories without constraint violation and has the smallest costs among the methods with zero violation. The max and mean constraint violations are computed for the violations accumulated throughout the trajectories. Better values are colored greener. Standard deviations over 5 runs are shown in parentheses.}
    \label{tab:cbf}
    \small
    \setlength{\tabcolsep}{3pt}
    \begin{tabular}{lccccc}
        \toprule
        {} & Cost & max violation & mean violation & $T_\textup{test}$ (ms) & $T_\textup{train}$ (min)\\
        \midrule
        CBF-QP & 
        \cellcolor[RGB]{237,204,211} 948.32 (0.00) & 
        \cellcolor[RGB]{216,238,221} 0.00 (0.00) & \cellcolor[RGB]{216,238,221} 0.00 (0.00) & \cellcolor[RGB]{237,204,211} 579.29 (4.60) & - \\
        NN & 
        \cellcolor[RGB]{216,238,221} 421.92 (0.15)& 
        \cellcolor[RGB]{237,204,211} 157.62 (0.90) & \cellcolor[RGB]{237,204,211} 118.92 (0.55) & \cellcolor[RGB]{205,227,216} 0.23 (0.01) & \cellcolor[RGB]{207,230,217} 256.34 (4.01)\\
        Soft &  
        \cellcolor[RGB]{254,250,235} 480.10 (0.54) &
        \cellcolor[RGB]{251,222,215} 6.92 (0.05)	& \cellcolor[RGB]{254,242,226} 3.95 (0.10) & \cellcolor[RGB]{206,228,217} 0.22 (0.00) & 
        \cellcolor[RGB]{207,230,217} 255.07 (0.77)\\
        DC3  &  
        \cellcolor[RGB]{254,250,235} 480.21 (0.71) & 
        \cellcolor[RGB]{251,222,215} 6.86 (0.13) & 
        \cellcolor[RGB]{254,242,226} 3.88 (0.12) & 
        \cellcolor[RGB]{249,219,214} 15.71 (0.29) & 
        \cellcolor[RGB]{242,208,211} 637.69 (6.71) \\
        \methodaff{}  &  
        \cellcolor[RGB]{252,225,217} 518.85 (8.71)	 & \cellcolor[RGB]{216,238,221} 0.00 (0.00) & \cellcolor[RGB]{216,238,221} 0.00 (0.00) & 
        \cellcolor[RGB]{217,238,222} 2.71 (0.06) & \cellcolor[RGB]{220,240,222} 370.05 (3.42) \\
        \bottomrule
    \end{tabular}
\end{table}

In this experiment, we apply \methodaff{} to enforce safety constraints in control systems. Consider a control-affine system with its known dynamics $f$ and $g$:
$\dot{x}(t) = f(x(t))+g(x(t)) u(t)$,
where $x(t)\in\mathbb{R}^{n_\textup{in}}$ is the system state, and $u(t)\in\mathbb{R}^{n_\textup{out}}$ is the control input at time $t$. 
For safety reasons (e.g., avoiding obstacles), the system requires $x(t)\in\mathcal{X}_\textup{safe}\subset\mathbb{R}^{n_\textup{in}}$ for all $t$. We translate this safety condition into a state-dependent affine constraint on the control input using a control barrier function (CBF) $h:\mathbb{R}^{n_\textup{in}}\rightarrow\mathbb{R}$~\citep{ames2019control}. Suppose its super-level set $\{x\in\mathbb{R}^{n_\textup{in}}|h(x)\geq 0\}\subset\mathcal{X}_\textup{safe}$ and $h(x(0))\geq0$.
Then, we can ensure $h(x(t))\geq0 \;\forall t\geq0$ by guaranteeing
\begin{equation}\label{eq:cbf_constraint}
    \dot{h}(x)=\!\nabla h(x)^\top\! \big(f(x)\!+\!g(x)\pi(x)\big)\geq\!-\alpha h(x)
\end{equation}
at each $x(t)$ for a state-feedback control policy $\pi:\mathbb{R}^{n_\textup{in}}\rightarrow\mathbb{R}^{n_\textup{out}}$ with some $\alpha>0$. 
Enforcing~\eqref{eq:cbf_constraint} for multiple CBFs ensures the trajectory remains within the intersection of the corresponding safe sets.

We consider controlling a unicycle system to minimize the cost over trajectories while avoiding collisions with two elliptical obstacles, each presented with a CBF (see Appendix~\ref{appendix:cbf} for details). Then, we can formulate the problem as an optimization problem with its objective function being the expected cost over the trajectory $x(t)$ generated by a parameterized state feedback policy $\pi_\theta:\mathbb{R}^{n_\textup{in}}\rightarrow\mathbb{R}^{n_\textup{out}}$ from random initial point $x(0)\sim\mathcal{D}$:
\begin{equation}\label{eq:traj_opt_cont}
\begin{aligned}
    \argmin_\theta\; \underset{x(0)\sim\mathcal{D}}{\mathbb{E}} \int_{t=0}^{T} [x^\top Q x + \pi_\theta(x)^\top R \pi_\theta(x)] dt
    \;\textup{ s.t. } \text{\eqref{eq:cbf_constraint} holds for all CBFs} \;\forall t\in[0,T]\; \forall i
\end{aligned}
\end{equation}
where $Q$ is the state cost matrix and $R$ is the control cost matrix.

Given a nominal controller $\pi_\textup{nom}\!:\mathbb{R}^{n_\textup{in}}\!\rightarrow\!\mathbb{R}^{n_\textup{out}}$ designed without considering obstacles, a conventional approach to find a safe controller is to solve the following quadratic program at each $x(t)$:
\begin{equation*}
\textup{CBF-QP}:
     \pi_\textup{CBF-QP}(x)=\argmin_u  \|u-\pi_\textup{nom}(x)\|_2
    \;\textup{ s.t. } \text{\eqref{eq:cbf_constraint} holds for all CBFs}.
\end{equation*}
The downside of this method is that the controller cannot optimize a cost/reward over trajectories, as it only stays close to the nominal controller. Instead, we can do so by training neural network policies $\pi_\theta(x):=\pi_\textup{nom}(x)+f_\theta(x)$ with neural networks $f_\theta$. For computation, we approximate \eqref{eq:traj_opt_cont} by minimizing the costs of rolled-out trajectories from randomly sampled initial states.
As shown in Fig.~\ref{fig:exp_cbf} and Table~\ref{tab:cbf}, \methodaff{} consistently guarantees safe trajectories with low costs.

\section{Conclusion}
\label{section:conclusion}

In this paper, we presented \method{}, a practical framework for constructing neural networks that inherently satisfy input--output constraints. We proved that imposing these hard constraints does not limit the expressive power of these neural networks by providing universal approximation guarantees. We demonstrated the utility and versatility of our method across several applications, such as learning with piecewise constraints, learning optimization solvers with guaranteed feasibility, and optimizing control policies in safety-critical systems.
Using \method{} in other application domains that benefit from incorporating domain-specific knowledge is a promising direction for future work. One such example is in \cite{tang2024learning}. Also, we aim to explore developing methods for performing fast projections for problems with more general constraints. Lastly, extending our approach to support other forms of inductive biases, such as equivariances and invariances, would potentially be of great interest.

\subsubsection*{Acknowledgments}
The authors thank Anoopkumar Sonar for his help during the early development of this manuscript.
The authors acknowledge the MIT SuperCloud and Lincoln Laboratory Supercomputing Center for providing computing resources that have contributed to the results reported within this paper. This work was supported in part by MathWorks, the MIT-IBM Watson AI Lab, the MIT-Amazon Science Hub, and the MIT-Google Program for Computing Innovation.

\vskip 0.2in
\bibliography{ref}

\newpage
\appendix

\section{Proofs}

\subsection{Proof of Proposition~\ref{prop:affine_opt}}
\label{appendix:affine_opt}

\AffOpt*
\begin{proof}
We first construct a minimum $\ell^2$-norm optimization that characterizes \methodaff{} based on a property of the pseudoinverse. Then, we show the optimization is equivalent to~\eqref{eq:affine_opt}.
We drop the input dependency on $x$ when it is evident to simplify the presentation.
Rewriting the closed-form projection of \methodaff{} in~\eqref{eq:hardnet_aff},
\begin{equation*}
    \mathcal{P}(f_\theta)(x)-f_\theta(x) = A^+ \big[\relu{}\big(b^l-Af_\theta(x)\big) - \relu{}\big(Af_\theta(x)-b^u\big)\big].
\end{equation*}
Based on the property of the pseudoinverse $A^+$ of a full row rank matrix $A$ that $z^*=A^+y$ finds the minimum $\ell^2$-norm solution among all solutions of the (under-determined) linear system $y=Az$, we have
\begin{equation*}
    \mathcal{P}(f_\theta)(x)\!-\!f_\theta(x) = \argmin_{z\in\mathbb{R}^{n_\textup{out}}} \|z\|_2
    \;\textup{ s.t. }
    Az = \relu{}\big(b^l\!-\! Af_\theta(x)\big) - \relu{}\big(Af_\theta(x)\!-\!b^u\big),
\end{equation*}
which implies 
\begin{equation*}
    \mathcal{P}(f_\theta)(x) = \displaystyle\argmin_{z\in\mathbb{R}^{n_\textup{out}}} \|z-f_\theta(x)\|_2 \;\textup{ s.t. } z\in\mathcal{Z}_\textup{eq}
\end{equation*}
with the feasible set
$\mathcal{Z}_\textup{eq}:=
\big\{z\in\mathbb{R}^{n_\textup{out}}| Az = Af_\theta(x) + \relu{}\big(b^l\!-\!Af_\theta(x)\big) - \relu{}\big(Af_\theta(x)\!-\!b^u\big)\big\}$.

Now, we prove the equation~\eqref{eq:affine_opt} by showing that the feasible set $\mathcal{Z}_\textup{eq}$ is the same as that of the outer optimization in~\eqref{eq:affine_opt}, denoted by $\mathcal{Z}_\textup{opt}:= \Big\{\displaystyle\argmin_{z\in\mathbb{R}^{n_\textup{out}}}
    \big\|A\big(z- f_\theta(x)\big)\big\|_2 
    \textup{ s.t. } b^l \leq A z \leq b^u\Big\}$.
Let $a_i\in\mathbb{R}^{n_\textup{c}}$ denote the $i$-th row of $A$. We first observe that, for any $z\in\mathcal{Z}_\textup{opt}$, 
\begin{align}
    \big\|A\big(z- f_\theta(x)\big)\big\|_2^2
    &= \sum_{i=1}^{n_c}(a_i^\top(z-f_\theta(x)))^2\\
    &\geq \sum_{i=1}^{n_c} \Big(\relu{}\big(b_{(i)}^l\!-\!a_i^\top f_\theta(x)\big) - \relu{}\big(a_i^\top f_\theta(x)\!-\!b_{(i)}^u\big)\Big)^2
    \label{eq:affine_opt_lb}\\
    &= \big\|\relu{}\big(b^l\!-\!Af_\theta(x)\big) - \relu{}\big(Af_\theta(x)\!-\!b^u\big)\big\|_2^2,
    \label{eq:affine_opt_lb2}
\end{align}
where the inequality~\eqref{eq:affine_opt_lb} holds for each summand. It can be easily verified by considering the component-wise constraint $b_{(i)}^l \leq a_i^\top z\leq b_{(i)}^u$ in three cases: \romannumeral 1) $a_i^\top f_\theta(x)<b_{(i)}^l$, \romannumeral 2) $b_{(i)}^l \leq a_i^\top f_\theta(x)\leq b_{(i)}^u$, and \romannumeral 3) $a_i^\top f_\theta(x)>b_{(i)}^u$.
With this observation, we show $\mathcal{Z}_\textup{eq}=\mathcal{Z}_\textup{opt}$:

\romannumeral 1) ($\mathcal{Z}_\textup{eq}\subset\mathcal{Z}_\textup{opt}$)
Suppose $z'\in\mathcal{Z}_\textup{eq}$. Then, following the same proof of Proposition~\ref{prop:affine_prop} in Appendix~\ref{appendix:affine_prop} with replacing $\mathcal{P}(f_\theta)(x)$ as $z'$, we can show $b^l \leq A z' \leq b^u$.
In other words, $z'$ satisfies the constraints of $\mathcal{Z}_\textup{opt}$. Also, $z'$ attains the lower bound of the objective of $\mathcal{Z}_\textup{opt}$ in~\eqref{eq:affine_opt_lb} since $A\big(z'-f_\theta(x)\big) = \relu{}\big(b^l\!-\!Af_\theta(x)\big) - \relu{}\big(Af_\theta(x)\!-\!b^u\big)$. This implies $z'\in\mathcal{Z}_\textup{opt}$.
Thus, $\mathcal{Z}_\textup{eq}\subset\mathcal{Z}_\textup{opt}$.

\romannumeral 2) ($\mathcal{Z}_\textup{opt}\subset\mathcal{Z}_\textup{eq}$)
Suppose $z'\in\mathcal{Z}_\textup{opt}$. Since $\mathcal{P}(f_\theta)(x)\in\mathcal{Z}_\textup{eq}\subset\mathcal{Z}_\textup{opt}$ attains the lower bound~\eqref{eq:affine_opt_lb2}, the inequality~\eqref{eq:affine_opt_lb} should hold with equality for each summand for any $z\in\mathcal{Z}_\textup{opt}$. In addition to the equality, if we consider the component-wise constraint $b_{(i)}^l \leq a_i^\top z\leq b_{(i)}^u$ in three cases: \romannumeral 1) $a_i^\top f_\theta(x)<b_{(i)}^l$, \romannumeral 2) $b_{(i)}^l \leq a_i^\top f_\theta(x)\leq b_{(i)}^u$, and \romannumeral 3) $a_i^\top f_\theta(x)>b_{(i)}^u$, we can show
\begin{equation*}
    A\big(z'-f_\theta(x)\big) = \relu{}\big(b^l\!-\!Af_\theta(x)\big) - \relu{}\big(Af_\theta(x)\!-\!b^u\big).
\end{equation*}
This implies $z'\in\mathcal{Z}_\textup{eq}$.
Thus, $\mathcal{Z}_\textup{opt}\subset\mathcal{Z}_\textup{eq}$.
\end{proof}

\subsection{Proof of Proposition~\ref{prop:affine_prop}}
\label{appendix:affine_prop}

\AffProp*
\begin{proof}
Given row index $i\in\{1,\cdots,n_\textup{c}\}$ and input $x\in\mathcal{X}$, for the plain neural network $f_\theta$, we can consider three cases: \romannumeral 1) $b^l_{(i)}(x)\leq a_i^\top f_\theta(x)\leq b^u_{(i)}(x)$, \romannumeral 2) $a_i^\top f_\theta(x)<b^l_{(i)}(x)$, and \romannumeral 3) $a_i^\top f_\theta(x)>b^u_{(i)}(x)$. In each case, we show that the projected output $\mathcal{P}(f_\theta)(x)$ satisfies the corresponding equality in Proposition~\ref{prop:affine_prop}.

\romannumeral 1) Suppose $b^l_{(i)}(x)\leq A(x)f_\theta(x)\leq b^u_{(i)}(x)$. From \eqref{eq:hardnet_aff},
\begin{equation*}
    A(x)\mathcal{P}(f_\theta)(x) = 
    A(x)f_\theta(x)+ \relu\big(b^l(x) - A(x)f_\theta(x)\big) - \relu\big(A(x)f_\theta(x)-b^u(x)\big).
\end{equation*}
Then, for the $i$-th component,
\begin{align*}
    a_i^\top\mathcal{P}(f_\theta)(x) =
    a_i^\top f_\theta(x)+ \relu\big(b^l_{(i)}(x) \!-\! a_i^\top f_\theta(x)\big) - \relu\big(a_i^\top f_\theta(x) \!-\! b^u_{(i)}(x)\big)
    = a_i^\top f_\theta(x)
\end{align*}
as $b^l_{(i)}(x) - a_i^\top f_\theta(x)\leq0$ and $a_i^\top f_\theta(x)-b^u_{(i)}(x)\leq0$.

\romannumeral 2) Suppose $a_i^\top f_\theta(x)<b^l_{(i)}(x)$. Then,
\begin{align*}
    a_i^\top\mathcal{P}(f_\theta)(x) &=
    a_i^\top f_\theta(x)+ \relu\big(b^l_{(i)}(x) - a_i^\top f_\theta(x)\big) - \relu\big(a_i^\top f_\theta(x)-b^u_{(i)}(x)\big)\\
    &= a_i^\top f_\theta(x) + \big(b^l_{(i)}(x) - a_i^\top f_\theta(x)\big) = b^l_{(i)}(x)
\end{align*}
as $b^l_{(i)}(x) - a_i^\top f_\theta(x)>0$ and $a_i^\top f_\theta(x)-b^u_{(i)}(x)<0$.

\romannumeral 3) Suppose $a_i^\top f_\theta(x)>b^u_{(i)}(x)$. Then,
\begin{align*}
    a_i^\top\mathcal{P}(f_\theta)(x) &=
    a_i^\top f_\theta(x)+ \relu\big(b^l_{(i)}(x) - a_i^\top f_\theta(x)\big) - \relu\big(a_i^\top f_\theta(x)-b^u_{(i)}(x)\big)\\
    &= a_i^\top f_\theta(x) - \big(a_i^\top f_\theta(x)-b^u_{(i)}(x)\big) = b^u_{(i)}(x)
\end{align*}
as $b^l_{(i)}(x) - a_i^\top f_\theta(x)<0$ and $a_i^\top f_\theta(x)-b^u_{(i)}(x)>0$.
This shows Proposition~\ref{prop:affine_prop} holds.
\end{proof}

\subsection{Proof of Theorem~\ref{thm:affine_uat}}
\label{appendix:affine_uat}

\AffUAT*
\begin{proof}
For any functions $f_\textup{NN}\in\mathcal{F}_\textup{NN}$ and $f_t\in\mathcal{F}_\textup{target}$, we first show that $\|f_t(x)-\mathcal{P}(f_\textup{NN})(x)\|_2$ can be bounded by some constant times $\|f_t(x)-f_\textup{NN}(x)\|_2$ for all $x\in\mathcal{X}$. We drop the input dependency on $x$ when it is evident to simplify the presentation. From \methodaff{} in \eqref{eq:hardnet_aff},
\begin{align}
    \|f_t(x)\!-\!\mathcal{P}(f_\textup{NN})(x)\|_2
    &= \big\|f_t(x) \!-\! f_\textup{NN}(x) \!-\! A^+ \big[\relu{}\big(b^l\!-\!Af_\textup{NN}(x)\big) \!-\! \relu{}\big(Af_\textup{NN}(x)\!-\!b^u\big)\big]\big\|_2\\
    &\hspace{-1.9cm}\leq \|f_t(x) - f_\textup{NN}(x)\|_2 + \big\|A^+ \relu{}\big(b^l\!-\!Af_\textup{NN}(x)\big)\big\|_2 + \big\|A^+ \relu{}\big(Af_\textup{NN}(x)\!-\!b^u\big)\big\|_2,
    \label{eq:diff_triangle}
\end{align}
by the triangle inequalities. Then, for the second term of RHS,
\begin{align}
    \big\|A^+ \relu{}\big(b^l-Af_\textup{NN}(x)\big)\big\|_2 &=
    \big\|A^+ \relu{}\big(b^l-Af_t(x)+A(f_t(x)-f_\textup{NN}(x))\big)\big\|_2\\
    &\leq \|A^+\|_2 \|\relu{}\big(b^l-Af_t(x)+A(f_t(x)-f_\textup{NN}(x))\big)\|_2\\
    &\leq \|A^+\|_2 \|A(f_t(x)-f_\textup{NN}(x))\|_2 \label{eq:relu_sum}\\
    &\leq \|A^+\|_2 \|A\|_2 \|f_t(x)-f_\textup{NN}(x)\|_2,
\end{align}
where we use the following lemma for \eqref{eq:relu_sum}:
\begin{lemma}
    For any $v,w\in\mathbb{R}^{n_\textup{c}}$, if $v\leq0$,
    $\|\relu{}(v+w)\|_2\leq\|w\|_2$.
\end{lemma}
For its proof, we consider a simpler case first. For $v,w\in\mathbb{R}$, if $v\leq0$, then $|\relu{}(v+w)|\leq |w|$. This is because if $w\leq -v\leq 0$, $|\relu{}(v+w)|=0\leq|w|$, else $w>-v\geq0$ so that $|\relu{}(v+w)|=v+w\leq w=|w|$.
Then, for $v,w\in\mathbb{R}^{n_\textup{c}}$, if $v\leq0$,
\begin{equation*}
    \|\relu{}(v+w)\|_2^2=\sum_{i=1}^{n_\textup{c}} |\relu{}(v_{(i)}+w_{(i)})|^2
    \leq \sum_{i=1}^{n_\textup{c}} |w_{(i)}|^2=\|w\|_2^2.
\end{equation*}
Thus, the lemma holds. This lemma implies \eqref{eq:relu_sum} as $b^l-Af_t(x)\leq 0$ since $f_t\in\mathcal{F}_\textup{target}$ satisfies the constraints~\eqref{eq:constraint_affine_multi}.
Similarly,
\begin{equation*}
    \big\|A^+ \relu{}\big(Af_\textup{NN}(x)-b^u\big)\big\|_2
    \leq \|A^+\|_2 \|A\|_2 \|f_\textup{NN}(x)-f_t(x)\|_2.
\end{equation*}
Then, putting them together in~\eqref{eq:diff_triangle}, we obtain
\begin{equation*}
    \|f_t(x)-\mathcal{P}(f_\textup{NN})(x)\|_2 \leq 
    \big(1 + 2\|A^+\|_2\|A\|_2\big)
    \|f_t(x)-f_\textup{NN}(x)\|_2.
\end{equation*}
Since $A(x)$ is continuous over the compact domain $\mathcal{X}$, there exists some constant $K>0$ such that
\begin{equation*}
    \big(1 + 2\|A^+\|_2\|A\|_2\big) \leq K,
\end{equation*}
for all $x\in\mathcal{X}$. Thus,
\begin{equation*}
    \|f_t(x)-\mathcal{P}(f_\textup{NN})(x)\|_2
    \leq K \|f_t(x)-f_\textup{NN}(x)\|_2.
\end{equation*}
Now, we extend this inequality to the general $\ell^p$-norm and $\ell^\infty$-norm by using the inequalities $\|v\|_q\leq\|v\|_r\leq m^{\frac{1}{r}-\frac{1}{q}}\|v\|_q$ for any $v\in\mathbb{R}^m$ and $q\geq r\geq 1$. If $p\leq2$,
\begin{align*}
    \|f_t(x)-\mathcal{P}(f_\textup{NN})(x)\|_p 
    &\leq n_\textup{out}^{\frac{1}{p}-\frac{1}{2}}\|f_t(x)-\mathcal{P}(f_\textup{NN})(x)\|_2\\
    &\leq n_\textup{out}^{\frac{1}{p}-\frac{1}{2}} K \|f_t(x)-f_\textup{NN}(x)\|_2
    \leq n_\textup{out}^{\frac{1}{p}-\frac{1}{2}} K \|f_t(x)-f_\textup{NN}(x)\|_p.
\end{align*}
If $p>2$,
\begin{align*}
    \|f_t(x)-\mathcal{P}(f_\textup{NN})(x)\|_p
    &\leq \|f_t(x)-\mathcal{P}(f_\textup{NN})(x)\|_2\\
    &\leq K \|f_t(x)-f_\textup{NN}(x)\|_2
    \leq K n_\textup{out}^{\frac{1}{2}-\frac{1}{p}}\|f_t(x)-f_\textup{NN}(x)\|_p.
\end{align*}
Thus, for any $p\in[1,\infty)$,
\begin{equation}\label{eq:ineq_p}
    \|f_t(x)-\mathcal{P}(f_\textup{NN})(x)\|_p
    \leq n_\textup{out}^{|\frac{1}{p}-\frac{1}{2}|} K \|f_t(x)-f_\textup{NN}(x)\|_p.
\end{equation}
Then, with $p\rightarrow\infty$,
\begin{equation}\label{eq:ineq_infty}
    \|f_t(x)-\mathcal{P}(f_\textup{NN})(x)\|_\infty \leq
    n_\textup{out}^\frac{1}{2} K \|f_t(x)-f_\textup{NN}(x)\|_\infty.
\end{equation}

Now, we prove the theorem. Suppose $\mathcal{F}_\textup{NN}, \mathcal{F} \subset \mathcal{C}(\mathcal{X},\mathbb{R}^{n_\textup{out}})$.
For any $f_t\in\mathcal{F}_\textup{target}\subset\mathcal{F}$ and $\epsilon>0$, we have a function $f_\textup{NN}\in\mathcal{F}_\textup{NN}$ such that 
\begin{equation}\label{eq:f_nn_infty}
    \|f_t-f_\textup{NN}\|_\infty \leq \dfrac{\epsilon}{n_\textup{out}^\frac{1}{2} K},
\end{equation}
since $\mathcal{F}_\textup{NN}$ universally approximates $\mathcal{F}$. For such $f_\textup{NN}$,
\begin{align*}
    \|f_t-\mathcal{P}(f_\textup{NN})\|_\infty 
    &= \sup_{x\in\mathcal{X}} \|f_t(x)-\mathcal{P}(f_\textup{NN})(x)\|_\infty\\
    &\leq \sup_{x\in\mathcal{X}} n_\textup{out}^\frac{1}{2} K \|f_t(x)-f_\textup{NN}(x)\|_\infty\\
    &= n_\textup{out}^\frac{1}{2} K \sup_{x\in\mathcal{X}} \|f_t(x)-f_\textup{NN}(x)\|_\infty
    = n_\textup{out}^\frac{1}{2} K \|f_t-f_\textup{NN}\|_\infty \leq \epsilon,
\end{align*}
where the first and last inequalities are from \eqref{eq:ineq_infty} and \eqref{eq:f_nn_infty}, respectively. Since $\mathcal{P}(f_\textup{NN})\in\mathcal{F}_\textup{\methodaff{}}$, $\mathcal{F}_\textup{\methodaff{}}$ universally approximates $\mathcal{F}_\textup{target}$.

Similarly, suppose $\mathcal{F}_\textup{NN}, \mathcal{F} \subset L^p(\mathcal{X},\mathbb{R}^{n_\textup{out}})$.
For any $f_t\in\mathcal{F}_\textup{target}\subset\mathcal{F}$ and $\epsilon>0$, we have a function $f_\textup{NN}\in\mathcal{F}_\textup{NN}$ such that 
\begin{equation}\label{eq:f_nn_p}
    \|f_t-f_\textup{NN}\|_p \leq \dfrac{\epsilon}{n_\textup{out}^{|\frac{1}{p}-\frac{1}{2}|} K},
\end{equation}
since $\mathcal{F}_\textup{NN}$ universally approximates $\mathcal{F}$. For such $f_\textup{NN}$,
\begin{align*}
    \|f_t-\mathcal{P}(f_\textup{NN})\|_p 
    &= \big(\int_{\mathcal{X}} \|f_t(x)-\mathcal{P}(f_\textup{NN})(x)\|_p^p dx\big)^{1/p}\\
    &\leq \big(\int_{\mathcal{X}} (n_\textup{out}^{|\frac{1}{p}-\frac{1}{2}|} K)^p \|f_t(x)-f_\textup{NN}(x)\|_p^p dx\big)^{1/p}\\
    &= n_\textup{out}^{|\frac{1}{p}-\frac{1}{2}|} K \big(\int_{\mathcal{X}} \|f_t(x)-f_\textup{NN}(x)\|_p^p dx\big)^{1/p}
    = n_\textup{out}^{|\frac{1}{p}-\frac{1}{2}|} K \|f_t-f_\textup{NN}\|_p \leq \epsilon,
\end{align*}
where the first and last inequalities are from \eqref{eq:ineq_p} and \eqref{eq:f_nn_p}, respectively. Since $\mathcal{P}(f_\textup{NN})\in\mathcal{F}_\textup{\methodaff{}}$, $\mathcal{F}_\textup{\methodaff{}}$ universally approximates $\mathcal{F}_\textup{target}$.
\end{proof}

\subsection{Proof of Theorem~\ref{thm:convex_uat}}
\label{appendix:convex_uat}

\CvxUAT*

As for \methodaff{}, we prove this theorem as a corollary of the following theorem:
\begin{theorem}
    Consider input-dependent sets $\mathcal{C}(x)\!\subset\!\mathbb{R}^{n_\textup{out}}$ that are convex for all $x\in\mathcal{X}\subset\mathbb{R}^{n_\textup{in}}$.
    Then, for any function classes $\mathcal{F}_\textup{NN}, \mathcal{F} \subset \mathcal{C}(\mathcal{X},\mathbb{R}^{n_\textup{out}})$ (or $\mathcal{F}_\textup{NN}, \mathcal{F} \subset L^p(\mathcal{X},\mathbb{R}^{n_\textup{out}})$ for any $p\in[1,\infty)$) and the projection $\mathcal{P}$ of \methodcvx{} in~\eqref{eq:hardnet_cvx},
    if $\mathcal{F}_\textup{NN}$ universally approximates $\mathcal{F}$,
    $\mathcal{F}_\textup{\methodcvx{}}:=\{\mathcal{P}(f_\textup{NN})| f_\textup{NN}\in\mathcal{F}_\textup{NN}\}$ universally approximates $\mathcal{F}_\textup{target}:=\{f_t\in\mathcal{F}| f_t(x)\in\mathcal{C}(x) \;\forall x\in\mathcal{X}\}$.
\end{theorem}

\begin{proof}
Similarly to the proof of Theorem~\ref{thm:affine_uat} in Appendix~\ref{appendix:affine_uat}, for any functions $f_\textup{NN}\in\mathcal{F}_\textup{NN}$ and $f_t\in\mathcal{F}_\textup{target}$, we first prove the following inequality:
\begin{equation*}
    \|f_t(x)-\mathcal{P}(f_\textup{NN})(x)\|_2 \leq \|f(x)-f_\textup{NN}(x)\|_2 \;\;\forall x\in\mathcal{X}.
\end{equation*}

Given $x\in\mathcal{X}$, consider the simple case $f_\textup{NN}(x)\in\mathcal{C}(x)$ first. Then, $\mathcal{P}(f_\textup{NN})(x)=f_\textup{NN}(x)$ from the projection in~\eqref{eq:hardnet_cvx} which satisfies $\|f_t(x)-\mathcal{P}(f_\textup{NN})(x)\|_2 \leq \|f_t(x)-f_\textup{NN}(x)\|_2$.

On the other hand, if $f_\textup{NN}(x)\notin\mathcal{C}(x)$,
consider the triangle connecting $f_\textup{NN}(x), \mathcal{P}(f_\textup{NN})(x)$ and $f_t(x)$. Then, the side between $f_\textup{NN}(x)$ and $\mathcal{P}(f_\textup{NN})(x)$ is orthogonal to the tangent hyperplane of the convex set $\mathcal{C}(x)$ at $\mathcal{P}(f_\textup{NN})(x)$. For the two half-spaces separated by the tangent hyperplane, $f_t(x)$ belongs to the other half-space than the one that contains $f_\textup{NN}(x)$ since $\mathcal{C}(x)$ is convex. Thus, the vertex angle at $\mathcal{P}(f_\textup{NN})(x)$ is larger than $\pi/2$. This implies that the side between $f_\textup{NN}(x)$ and $f_t(x)$ is the longest side of the triangle, so $\|f_t(x)-\mathcal{P}(f_\textup{NN})(x)\|_2 \leq \|f_t(x)-f_\textup{NN}(x)\|_2$.

Then, We can extend this $\ell^2$-norm result to general $\ell^p$-norm and $\ell^\infty$-norm as in Appendix~\ref{appendix:affine_uat}:
\begin{equation}\label{eq:ineq_p_cvx}
    \|f(x)-\mathcal{P}(f_\textup{NN})(x)\|_p \leq n_\textup{out}^{|\frac{1}{p}-\frac{1}{2}|} \|f(x)-f_\textup{NN}(x)\|_p.
\end{equation}
With $p\rightarrow\infty$,
\begin{equation}\label{eq:ineq_infty_cvx}
    \|f(x)-\mathcal{P}(f_\textup{NN})(x)\|_\infty \leq n_\textup{out}^{\frac{1}{2}} \|f(x)-f_\textup{NN}(x)\|_\infty.
\end{equation}

Now, we prove the theorem. Suppose $\mathcal{F}_\textup{NN}, \mathcal{F} \subset \mathcal{C}(\mathcal{X},\mathbb{R}^{n_\textup{out}})$.
For any $f_t\in\mathcal{F}_\textup{target}\subset\mathcal{F}$ and $\epsilon>0$, we have a function $f_\textup{NN}\in\mathcal{F}_\textup{NN}$ such that 
\begin{equation}\label{eq:f_nn_infty_cvx}
    \|f_t-f_\textup{NN}\|_\infty \leq \dfrac{\epsilon}{n_\textup{out}^\frac{1}{2}},
\end{equation}
since $\mathcal{F}_\textup{NN}$ universally approximates $\mathcal{F}$. For such $f_\textup{NN}$,
\begin{align*}
    \|f_t-\mathcal{P}(f_\textup{NN})\|_\infty 
    &= \sup_{x\in\mathcal{X}} \|f_t(x)-\mathcal{P}(f_\textup{NN})(x)\|_\infty\\
    &\leq \sup_{x\in\mathcal{X}} n_\textup{out}^\frac{1}{2} \|f_t(x)-f_\textup{NN}(x)\|_\infty\\
    &= n_\textup{out}^\frac{1}{2} \sup_{x\in\mathcal{X}} \|f_t(x)-f_\textup{NN}(x)\|_\infty
    = n_\textup{out}^\frac{1}{2} \|f_t-f_\textup{NN}\|_\infty \leq \epsilon,
\end{align*}
where the first and last inequalities are from \eqref{eq:ineq_infty_cvx} and \eqref{eq:f_nn_infty_cvx}, respectively. Since $\mathcal{P}(f_\textup{NN})\in\mathcal{F}_\textup{\methodcvx{}}$, $\mathcal{F}_\textup{\methodcvx{}}$ universally approximates $\mathcal{F}_\textup{target}$.

Similarly, suppose $\mathcal{F}_\textup{NN}, \mathcal{F} \subset L^p(\mathcal{X},\mathbb{R}^{n_\textup{out}})$.
For any $f_t\in\mathcal{F}_\textup{target}\subset\mathcal{F}$ and $\epsilon>0$, we have a function $f_\textup{NN}\in\mathcal{F}_\textup{NN}$ such that 
\begin{equation}\label{eq:f_nn_p_cvx}
    \|f_t-f_\textup{NN}\|_p \leq \dfrac{\epsilon}{n_\textup{out}^{|\frac{1}{p}-\frac{1}{2}|}},
\end{equation}
since $\mathcal{F}_\textup{NN}$ universally approximates $\mathcal{F}$. For such $f_\textup{NN}$,
\begin{align*}
    \|f_t-\mathcal{P}(f_\textup{NN})\|_p 
    &= \big(\int_{\mathcal{X}} \|f_t(x)-\mathcal{P}(f_\textup{NN})(x)\|_p^p dx\big)^{1/p}\\
    &\leq \big(\int_{\mathcal{X}} (n_\textup{out}^{|\frac{1}{p}-\frac{1}{2}|})^p \|f_t(x)-f_\textup{NN}(x)\|_p^p dx\big)^{1/p}\\
    &= n_\textup{out}^{|\frac{1}{p}-\frac{1}{2}|} \big(\int_{\mathcal{X}} \|f_t(x)-f_\textup{NN}(x)\|_p^p dx\big)^{1/p}
    = n_\textup{out}^{|\frac{1}{p}-\frac{1}{2}|} \|f_t-f_\textup{NN}\|_p \leq \epsilon,
\end{align*}
where the first and last inequalities are from \eqref{eq:ineq_p_cvx} and \eqref{eq:f_nn_p_cvx}, respectively. Since $\mathcal{P}(f_\textup{NN})\in\mathcal{F}_\textup{\methodcvx{}}$, $\mathcal{F}_\textup{\methodcvx{}}$ universally approximates $\mathcal{F}_\textup{target}$.
\end{proof}

Thus, by utilizing Theorem~\ref{thm:uat_deep} in this theorem, we have Theorem~\ref{thm:convex_uat} as a corollary.

\section{An Alternative Approach to Handle Equality Constraints Separately} \label{appendix:equality}

In this section, we propose an alternative approach to handle equality constraints separately, rather than setting $b^l=b^u$ in~\eqref{eq:constraint_affine_multi}. We first define some additional notation:
\paragraph{Additional notation} $v_{(i)}\in\mathbb{R}$, $v_{(:i)}\in\mathbb{R}^i$, and $v_{(i:)}\in\mathbb{R}^{m-i}$ denote the $i$-th component, the first $i$ and the last $m-i$ components of $v$, respectively. Similarly, $A_{(:i)}\in\mathbb{R}^{k\times i}$ and $A_{(i:)}\in\mathbb{R}^{k\times (m-i)}$ denote the first $i$ and the last $m-i$ columns of $A$, respectively. 

Now, suppose we have multiple input-dependent affine constraints in an aggregated form:
\begin{equation}\label{eq:constraint_eq_multi}
    A(x)f(x)\leq b(x),\;\; C(x)f(x)=d(x)\;\;\;\forall x\in\mathcal{X},
\end{equation}
where $A(x)\in\mathbb{R}^{n_\textup{ineq}\times n_\textup{out}}$, $b(x)\in\mathbb{R}^{n_\textup{ineq}}$, $C(x)\in\mathbb{R}^{n_\textup{eq}\times n_\textup{out}}$, $d(x)\in\mathbb{R}^{n_\textup{eq}}$ for $n_\textup{ineq}$ inequality and $n_\textup{eq}$ equality constraints.
For partitions $A(x)=[A_{(:n_\textup{eq})}\; A_{(n_\textup{eq}:)}]$ and $C(x)=[C_{(:n_\textup{eq})}\; C_{(n_\textup{eq}:)}]$, we make the following assumptions about the constraints:
\begin{assumption}\label{asmp:affine_eq}
    For each $x\in\mathcal{X}$,
    \romannumeral 1) there exists $y\in\mathbb{R}^{n_\textup{out}}$ that satisfies the constraints in~\eqref{eq:constraint_eq_multi},
    \;\romannumeral 2) $C_{(:n_\textup{eq})}$ is invertible, and
    \romannumeral 3) $\tilde{A}(x):=A_{(n_\textup{eq}:)}\!-\! A_{(:n_\textup{eq})}C_{(:n_\textup{eq})}^{-1}C_{(n_\textup{eq}:)}$ has full row rank.
\end{assumption}
Given $x\in\mathcal{X}$, when $C(x)$ has full row rank (i.e., no redundant constraints), there exists an invertible submatrix of $C(x)$ with its $n_\textup{eq}$ columns. Without loss of generality, we can assume $C_{(:n_\textup{eq})}$ is such submatrix by considering a proper permutation of the components of $f$. Then, the second assumption holds when the same permutation lets $C_{(:n_\textup{eq})}$ invertible for all $x\in\mathcal{X}$. The last assumption requires the total number of the constraints $n_\textup{ineq}+n_\textup{eq}$ to be less than or equal to the output dimension $n_\textup{out}$ and $A(x)$ to have full row rank.

Under the assumptions, we first efficiently reduce the $n_\textup{ineq}+n_\textup{eq}$ constraints to $n_\textup{ineq}$ equivalent inequality constraints on partial outputs $f_{(n_\textup{eq}:)}$ for a partition of the function $f(x)=[f_{(:n_\textup{eq})}; f_{(n_\textup{eq}:)}]$. Consider the hyperplane in the codomain $\mathcal{Y}$ over which the equality constraints are satisfied. Then, for the function output $f(x)$ to be on the hyperplane, the first part $f_{(:n_\textup{eq})}$ is determined by $f_{(n_\textup{eq}:)}$:
\begin{equation}\label{eq:f2_eq}
    f_{(:n_\textup{eq})}(x) = C_{(:n_\textup{eq})}^{-1} \big(d(x)-C_{(n_\textup{eq}:)} f_{(n_\textup{eq}:)}(x)\big).
\end{equation}
Substituting this $f_{(:n_\textup{eq})}$ into the inequality constraints, the constraints in \eqref{eq:constraint_eq_multi} is  equivalent to the following inequality constraints with \eqref{eq:f2_eq}:
\begin{equation*}
    \big(
    \underbrace{A_{(n_\textup{eq}:)} - A_{(:n_\textup{eq})} C_{(:n_\textup{eq})}^{-1} C_{(n_\textup{eq}:)}}
    _{=:\tilde{A}(x)}
    \big) 
    f_{(n_\textup{eq}:)}(x) \leq 
    \underbrace{b(x) - A_{(:n_\textup{eq})} C_{(:n_\textup{eq})}^{-1} d(x)}
    _{=:\tilde{b}(x)} 
    \;\forall x\in\mathcal{X}.
\end{equation*}

We propose a closed-form projection to enforce this $n_\textup{ineq}$ equivalent inequality constraints on $f_{(n_\textup{eq}:)}$, while the first part $f_{(:n_\textup{eq})}$ of the function is completely determined by the second part $f_{(n_\textup{eq}:)}$ as in~\eqref{eq:f2_eq}. We let the parameterized function $f_\theta:\mathcal{X}\rightarrow\mathbb{R}^{n_\textup{out}-n_\textup{eq}}$ approximate only the second part (or disregard the first $n_\textup{eq}$ outputs if $f_\theta(x)\in\mathbb{R}^{n_\textup{out}}$ is given). Then, we project $f_\theta$ to satisfy the constraints in~\eqref{eq:constraint_eq_multi} as below:
\begin{equation}\label{eq:proj_affine_multi}
    \mathcal{P}(f_\theta)(x)
    \!=\!\begin{bmatrix}
        C_{(:n_\textup{eq})}^{-1} \big(d(x)-C_{(n_\textup{eq}:)} f_\theta^*(x)\big)
        ;\;
        f_\theta^*(x)
    \end{bmatrix}
   ,
\end{equation}
where $f_\theta^*(x)
    := f_\theta(x)-\tilde{A}(x)^+ \relu{}\big(\tilde{A}(x)f_\theta(x)-\tilde{b}(x)\big)$ for all $x\in\mathcal{X}$
with $M^+:=M^\top (MM^\top)^{-1}$ denoting the pseudoinverse of a matrix $M$. This novel projection satisfies the following properties.
\begin{proposition}\label{prop:hardnet_eq_prop}
    Under Assumption~\ref{asmp:affine_eq}, for any parameterized (neural network) function $f_\theta:\mathcal{X}\rightarrow\mathbb{R}^{n_\textup{out}-n_\textup{eq}}$ and for all $x\in\mathcal{X}$, the projection $\mathcal{P}(f_\theta)$ in~\eqref{eq:proj_affine_multi} satisfies \vspace{.1cm}\\
    \romannumeral 1) $A(x)\mathcal{P}(f_\theta)(x)\leq b(x)$, \;\;
    \romannumeral 2) $C(x)\mathcal{P}(f_\theta)(x)=d(x)$,\\
    \romannumeral 3) For each $i$-th row $a_i\in\mathbb{R}^{n_\textup{out}}$ of $A(x)$,
    \hspace*{0.2cm}$a_i^\top \mathcal{P}(f_\theta)(x) = \begin{cases}
        a_i^\top \bar{f_\theta}(x)
        & \textup{if } \;a_i^\top \bar{f_\theta}(x)\leq b_{(i)}(x)\\
        b_{(i)}(x) & \textup{o.w.}
    \end{cases}$,\\
    where $\bar{f_\theta}(x):=
    \big[ 
    [C_{(:n_\textup{eq})}^{-1} \big(d(x)-C_{(n_\textup{eq}:)}f_\theta(x)\big)]; 
    \;f_\theta(x)
    \big]
    \in\mathbb{R}^{n_\textup{out}}$.
\end{proposition}
\begin{proof}
We simplify the notation of the partition by
$(\cdot)_1:= (\cdot)_{(:n_\textup{eq})}$ and $(\cdot)_2:=(\cdot)_{(n_\textup{eq}:)}$. Also, we drop the input dependency on $x$ when it is evident. Then,
\begin{align*}
    A(x)\mathcal{P}(f_\theta)(x)
    &= A_1 \mathcal{P}(f_\theta)_1 + A_2 \mathcal{P}(f_\theta)_2\\
    &= A_1 C_1^{-1}(d-C_2 f_\theta^*)
    + A_2 f_\theta^*\\
    &= (A_2 - A_1 C_1^{-1} C_2)f_\theta^* + A_1 C_1^{-1} d\\
    &= \tilde{A} f_\theta^*
    -\tilde{b} + b
    = \tilde{A} f_\theta
    - \relu{}(\tilde{A} f_\theta
    - \tilde{b}) -\tilde{b} + b
    \leq \tilde{b}-\tilde{b} + b = b(x).
\end{align*}
This shows (\romannumeral 1). For (\romannumeral 2),
\begin{equation*}
    C(x)\mathcal{P}(f_\theta)(x)
    = C_1 \mathcal{P}(f_\theta)_1 + C_2 \mathcal{P}(f_\theta)_2
    = (d-C_2 f_\theta^*) + C_2 f_\theta^* = d(x).
\end{equation*}
For (\romannumeral 3), we first observe that
\begin{equation*}
    a_i^\top \bar{f_\theta}(x)\leq b_{(i)}(x)
    \iff a_{i1}^\top f_\theta 
    + a_{i2}^\top C_2^{-1}(d-C_1 f_\theta) \leq b_{(i)}
    \iff \tilde{a}_i^\top f_\theta \leq \tilde{b}_{(i)}.
\end{equation*}
Then, if $a_i^\top \bar{f_\theta}(x)\leq b_{(i)}(x)$,
\begin{align*}
    a_i^\top \mathcal{P}(f_\theta)(x)
    &= a_{i1}^\top \mathcal{P}(f_\theta)_1 + a_{i2}^\top \mathcal{P}(f_\theta)_2\\
    &= (a_{i2}^\top - a_{i1}^\top C_1^{-1} C_2)f_\theta^* + a_{i1}^\top C_1^{-1} d
    = (a_{i2}^\top - a_{i1}^\top C_1^{-1} C_2)f_\theta + a_{i1}^\top C_1^{-1} d
    = a_i^\top \bar{f_\theta}(x), 
\end{align*}
where the second last equality is from $\tilde{A}f_\theta^* = \tilde{A}f_\theta - \relu{}(\tilde{A} f_\theta - \tilde{b})$ and $\tilde{a}_i^\top f_\theta \leq \tilde{b}_{(i)}$. Similarly, if $a_i^\top \bar{f_\theta}(x)> b_{(i)}(x)$, 
\begin{equation*}
    a_i^\top \mathcal{P}(f_\theta)(x)
    = (a_{i2}^\top - a_{i1}^\top C_1^{-1} C_2)f_\theta^* + a_{i1}^\top C_1^{-1} d
    = \tilde{b}_{(i)} + a_{i1}^\top C_1^{-1} d
    = b_{(i)}(x)
\end{equation*}

\end{proof}

While the projection~\eqref{eq:proj_affine_multi} guarantees the satisfaction of the hard constraints~\eqref{eq:constraint_eq_multi}, we can also rigorously show that it preserves the neural network's expressive power by the following universal approximation theorem, as in Appendix~\ref{appendix:affine_uat}.
\begin{theorem}\label{thm:eq_uat}
    Consider input-dependent constraints~\eqref{eq:constraint_eq_multi} that satisfy assumption~\ref{asmp:affine_eq}. Suppose $\mathcal{X}\subset\mathbb{R}^{n_\textup{in}}$ is compact, and $A(x), C(x)$ are continuous over $\mathcal{X}$.
    For any $p\in[1,\infty)$, let $\mathcal{F}=\{f\in L^p(\mathcal{X},\mathbb{R}^{n_\textup{out}})| f \text{ satisfies~\eqref{eq:constraint_affine_multi}}\}$.
    Then, the projection with $w$-width \textup{$\relu{}$} neural networks defined in~\eqref{eq:proj_affine_multi} \textit{universally approximates} $\mathcal{F}$ if $w\geq\max\{n_\textup{in}+1,n_\textup{out}-n_\textup{eq}\}$.
\end{theorem}
\begin{proof}
We first show that $\|f(x)-\mathcal{P}(f_\theta)(x)\|_2$ can be bounded by some constant times $\|f_{(n_{\textup{eq}}:)}(x)-f_\theta(x)\|_2$.
From \eqref{eq:proj_affine_multi},
\begin{align*}
    \|f(x)-\mathcal{P}(f_\theta)(x)\|_2^2
    &= \|f_{(:n_{\textup{eq}})}(x) - \mathcal{P}(f_\theta)_{(:n_{\textup{eq}})}(x)\|_2^2
    + \|f_{(n_{\textup{eq}}:)}(x) - \mathcal{P}(f_\theta)_{(n_{\textup{eq}}:)}(x)\|_2^2\\
    &= \|C_{(:n_\textup{eq})}^{-1}C_{(n_\textup{eq}:)}(f_{(n_{\textup{eq}:})}(x) - f_\theta^*(x))\|_2^2
    + \|f_{(n_{\textup{eq}}:)}(x) - f_\theta^*(x)\|_2^2\\
    &\leq \big(1 + \|C_{(:n_\textup{eq})}^{-1}C_{(n_\textup{eq}:)}\|_2^2\big) 
    \|f_{(n_{\textup{eq}}:)}(x) - f_\theta^*(x)\|_2^2,
\end{align*}
where the second equality holds by substituting $f_{(:n_\textup{eq})}$ in~\eqref{eq:f2_eq}. Meanwhile,
\begin{align*}
        \|f_{(n_{\textup{eq}}:)}(x) \!-\! f_\theta^*(x)\|_2 
        &\leq \|f_{(n_{\textup{eq}}:)}(x)\!-\!f_\theta(x)\|_2 
        + \|\tilde{A}^+\relu{}\big(\tilde{A}f_\theta(x)-\tilde{b}\big)\|_2\\
        &\leq \|f_{(n_{\textup{eq}}:)}(x) \!-\! f_\theta(x)\|_2 
        + \|\tilde{A}^+\|_2 \|\relu{}\big(\tilde{A}f_{(n_{\textup{eq}}:)} -\tilde{b} + \tilde{A}(f_\theta-f_{(n_{\textup{eq}}:)})\big)\|_2\\
        &\leq \|f_{(n_{\textup{eq}}:)}(x)\!-\!f_\theta(x)\|_2 
        + \|\tilde{A}^+\|_2 \|\tilde{A}(f_\theta(x)-f_{(n_{\textup{eq}}:)}(x))\|_2\\
        &\leq (1+\|\tilde{A}^+\|_2\|\tilde{A}\|_2) \|f_{(n_{\textup{eq}}:)}(x)-f_\theta(x)\|_2.
\end{align*}
Then, putting them together, we obtain
\begin{equation*}
    \|f(x)-\mathcal{P}(f_\theta)(x)\|_2 \leq 
    \big(1 + \|\tilde{A}^+\|_2\|\tilde{A}\|_2\big)
    \sqrt{(1+\|C_{(:n_\textup{eq})}^{-1}C_{(n_\textup{eq}:)}\|_2^2)}
    \|f_{(n_{\textup{eq}}:)}(x)-f_\theta(x)\|_2.
\end{equation*}
Since $A(x),C(x)$ are continuous over the compact domain $\mathcal{X}$, there exists some constant $K>0$ s.t.
\begin{equation*}
    \big(1 + \|\tilde{A}^+\|_2\|\tilde{A}\|_2\big)
    \sqrt{(1+\|C_{(:n_\textup{eq})}^{-1}C_{(n_\textup{eq}:)}\|_2^2)} \leq K
\end{equation*}
for all $x\in\mathcal{X}$. Thus,
\begin{equation*}
    \|f(x)-\mathcal{P}(f_\theta)(x)\|_2
    \leq K \|f_{(n_{\textup{eq}}:)}(x)-f_\theta(x)\|_2
\end{equation*}
Extending the inequalities to general $\ell^p$-norm for $p\geq 1$ by using the inequalities $\|v\|_q\leq\|v\|_r\leq m^{\frac{1}{r}-\frac{1}{q}}\|v\|_q$ for any $v\in\mathbb{R}^m$ and $q\geq r\geq 1$,
\begin{equation*}
    \|f(x)-\mathcal{P}(f_\theta)(x)\|_p
    \leq (n_\textup{out}-n_\textup{eq})^{|\frac{1}{p}-\frac{1}{2}|} K \|f_{(n_{\textup{eq}}:)}(x)-f_\theta(x)\|_p
\end{equation*}
Then, $f_\theta$ being dense in $L^p(\mathcal{X},\mathbb{R}^{n_\textup{out}-n_\textup{eq}})$ implies $\mathcal{P}(f_\theta)$ being dense in $L^p(\mathcal{X},\mathbb{R}^{n_\textup{out}})$. Thus, we can employ any universal approximation theorem for $f_\theta$ and convert it to that for $\mathcal{P}(f_\theta)$. While we utilize Theorem~\ref{thm:uat_deep} in this theorem, other universal approximation theorems on plain neural networks, such as Theorem~\ref{thm:uat_shallow}, can also be employed.
\end{proof}

\section{Gradient Properties of \methodaff{}}
\label{appendix:grad_aff}

This section investigates how the enforcement layer in \methodaff{} affects gradient computation. For simplicity, we focus on the case where the constraints are $A(x)f(x)\leq b(x) \;\forall x\in\mathcal{X}$.
For a datapoint $(x, y)$, consider the loss function $\ell\big(\mathcal{P}(f_\theta(x)), y\big)$. Using the chain rule, the gradient is given by:
\begin{equation}\label{eq:gradient}
    \nabla_\theta \ell\big(\mathcal{P}(f_\theta(x)), y\big)^\top 
    = \frac{\partial\ell\big(\mathcal{P}(f_\theta(x)), y\big)}{\partial\mathcal{P}(f_\theta(x))}
    \frac{\partial\mathcal{P}(f_\theta(x))}{\partial f_\theta(x)}
    \frac{\partial f_\theta(x)}{\partial\theta}.
\end{equation}
Here, the Jacobian of the enforcement layer $\frac{\partial\mathcal{P}(f_\theta(x))}{\partial f_\theta(x)}\in\mathbb{R}^{n_\textup{out}\times n_\textup{out}}$ plays a key role. Let $v_i:=\mathds{1}\{a_i(x)^\top f_\theta(x)>b_{(i)}(x)\}$ indicate whether the $i$-th constraint is violated by $f_\theta(x)$. Then,
\begin{equation*}
    \frac{\partial\mathcal{P}(f_\theta(x))}{\partial f_\theta(x)}
    = I - A^+\begin{bmatrix}
        v_1 a_1^\top\\
        \vdots\\
        v_{n_\textup{c}} a_{n_\textup{c}}^\top
    \end{bmatrix}.
\end{equation*}

Two critical properties of this Jacobian can lead to zero gradient in~\eqref{eq:gradient}. First, if the number of constraints equals the output dimension ($n_\textup{c}=n_\textup{out}$) and all constraints are violated ($v_i=1\;\forall i$), then the Jacobian becomes zero, causing the gradient~\eqref{eq:gradient} to vanish. Note that this issue never happens when $n_\textup{c}<n_\textup{out}$.
Second, for each $i\in\{1,2,\dots,n_\textup{c}\}$, the following holds:
\begin{equation*}
    a_i^\top\frac{\partial\mathcal{P}(f_\theta(x))}{\partial f_\theta(x)}
    = a_i^\top - a_i^\top A^+\begin{bmatrix}
        v_1 a_1^\top\\
        \vdots\\
        v_{n_\textup{c}} a_{n_\textup{c}}^\top
    \end{bmatrix}
    = a_i^\top - e_i^\top \begin{bmatrix}
        v_1 a_1^\top\\
        \vdots\\
        v_{n_\textup{c}} a_{n_\textup{c}}^\top
    \end{bmatrix}
    = a_i^\top - v_i a_i^\top.
\end{equation*}
This implies that if the loss gradient with respect to the projected output, $\big(\frac{\partial\ell\big(\mathcal{P}(f_\theta(x)), y\big)}{\partial\mathcal{P}(f_\theta(x))}\big)^\top\in\mathbb{R}^{n_\textup{out}}$, lies in the span of $\{a_i|i\in\{1,\dots,n_\textup{c}\}, v_i=1\}$, then the overall gradient~\eqref{eq:gradient} becomes zero. This case in fact subsumes the first case, as when $n_\textup{c}=n_\textup{out}$ and  $v_i=1\;\forall i$, the constraint vectors set spans the entire output space $\mathbb{R}^{n_\textup{out}}$.

\begin{figure}
    \centering
    \includegraphics[width=0.4\linewidth]{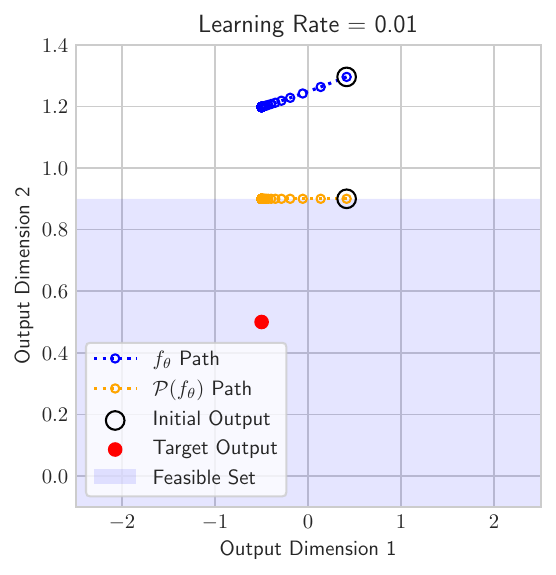}
    \includegraphics[width=0.4\linewidth]{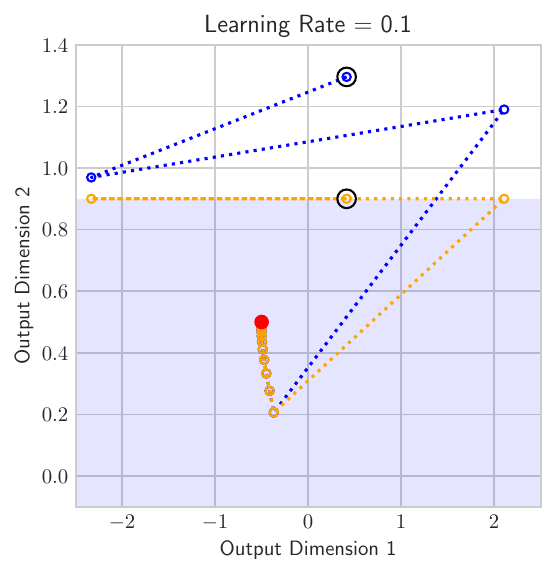}
    \includegraphics[width=0.4\linewidth]{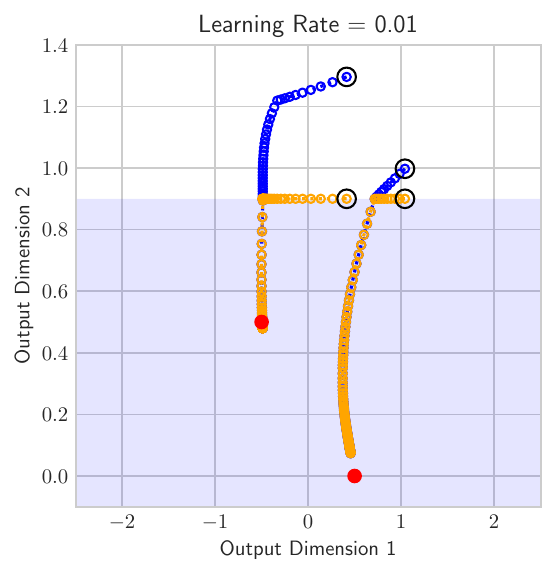}
    \includegraphics[width=0.4\linewidth]{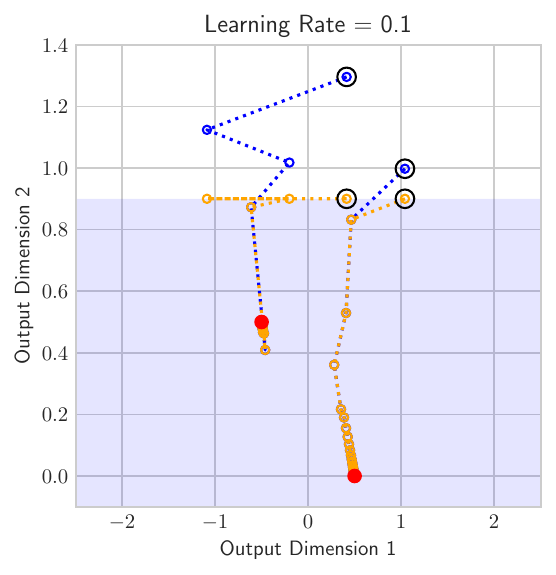}
    \caption{Visualization of 100 gradient descent steps for training a \methodaff{} model on a single datapoint (first row) and two datapoints (second row) from the same initialization, using two different learning rates (0.01 and 0.1). With the smaller learning rate, training on a single datapoint results in a zero gradient due to the enforcement layer (top left). However, when training on both datapoints, the vanishing gradient for the first datapoint is mitigated by the nonzero gradient from the second datapoint (bottom left). Also, using the larger learning rate enables the model to avoid the vanishing gradient issue, even when trained on the single datapoint (top right).}
    \label{fig:gd_path}
\end{figure}

However, such special cases are rare in practical settings, particularly when the model is trained on batched data. Even when zero gradients occur for certain datapoints, they can be offset by nonzero gradients from the other datapoints within the batch. This averaging effect allows the model to update in a direction that decreases the overall loss function. 
Thus, \methodaff{} allows training the projected function to achieve values (strictly) within the feasible set using conventional gradient-based algorithms. Additionally, one can promote the model $f_\theta$ to be initialized within the feasible set using the warm-start scheme outlined in Appendix~\ref{appendix:warmstart}, which involves training the model without the enforcement layer for a few initial epochs while regularizing with constraint violations.

We demonstrate the gradient behaviors discussed earlier using simple simulations of training \methodaff{} with the conventional gradient-descent algorithm. Consider two datapoints: $d_1=(-1, [-0.5, 0.5]^\top)$ and $d_2=(1, [0.5, 0]^\top)$, and a neural network $f_\theta:\mathbb{R}\rightarrow\mathbb{R}^2$ with two hidden layers, each containing 10 neurons with $\relu{}$ activations. The model enforces the input-independent constraint $[0, 1]\mathcal{P}(f_\theta)(x)\leq 0.9$ using \methodaff{}. Starting from the same initialization, the model is trained to minimize the squared error loss, first on $d_1$ alone and then on both datapoints, using two different learning rates (0.01 and 0.1), as shown in Fig.~\ref{fig:gd_path}.

Initially, $f_\theta$ violates the constraint on both datapoints. When trained on $d_1$ alone with a learning rate of 0.01, the optimization path converges to a point where the loss gradient with respect to the projected output is orthogonal to the gradient boundary, causing the overall gradient~\eqref{eq:gradient} to vanish. However, when the model is trained on both datapoints, the vanishing gradient for $d_1$ is mitigated by the nonzero gradient for $d_2$, enabling the model to achieve target values strictly within the feasible set. Furthermore, using a larger learning rate (0.1) allows the model to avoid the vanishing gradient issue and reach the target value even when trained solely on $d_1$.

\subsection{Learning with Warm Start}\label{appendix:warmstart}
In addition to the \method{} architecture shown in Figure~\ref{fig:schematic} that consists of a neural network $f_\theta$ and a differentiable enforcement layer with a projection $\mathcal{P}$ appended at the end, we propose a training scheme that can potentially result in better-optimized models. For the first $k$ epochs of training, we disable the enforcement layer and train the plain neural network $f_\theta$. Then, from the $(k\!+\!1)$-th epoch, we train on the projected model $\mathcal{P}(f_\theta)$. During the $k$ epochs of warm start, the neural network $f_\theta$ can be trained in a soft-constrained manner by regularizing the violations of constraints. In this paper, we train the \methodaff{} models without the warm-start scheme for simplicity, except in Section~\ref{section:exp_opt} where we use the warm-start for the initial 100 epochs. 

\section{Experimental Details}

\subsection{Details for the Learning with Piecewise Constraints Experiment}
\label{appendix:pwc}
The target function and constraints are as below:
\begin{equation*}
    f(x) \!=\! \begin{cases}
        -5\sin{\frac{\pi}{2}(x\!+\!1)}
        & \textup{if } x\leq -1\\
        0 & \textup{if } x\in(-1,0]\\
        4-9(x-\frac{2}{3})^2
        & \textup{if } x\in(0,1]\\
        5(1-x)+3 & \textup{if } x>1
    \end{cases}
    ,
    \textup{Constraints}\!:\! \begin{cases}
        f(x) \geq 5\sin^2{\frac{\pi}{2}(x+1)}
        & \textup{if } x\leq -1\\
        f(x) \leq 0 & \textup{if } x\in(-1,0]\\
        f(x) \geq \big(4\!-\!9(x\!-\!\frac{2}{3})^2\big)x
        & \textup{if } x\in(0,1]\\
        f(x) \leq 4.5(1-x)+3 & \textup{if } x>1
    \end{cases}.
\end{equation*}
These four constraints can be aggregated into the following single affine constraint: 
\begin{equation*}
    a(x)f(x)\leq b(x): \begin{cases}
        a(x)=-1, b(x)=-5\sin^2{\frac{\pi}{2}(x+1)}
        & \textup{if } x\leq -1\\
        a(x)=1, \;\;\;b(x)=0 & \textup{if } x\in(-1,0]\\
        a(x)=-1, b(x)=\big(9(x\!-\!\frac{2}{3})^2\!-\!4\big)x
        & \textup{if } x\in(0,1]\\
        a(x)=1, \;\;\;b(x)=4.5(1-x)+3 & \textup{if } x>1
    \end{cases}.
\end{equation*}

\begin{figure}[t]
    \centering
    \includegraphics[width=0.49\linewidth]{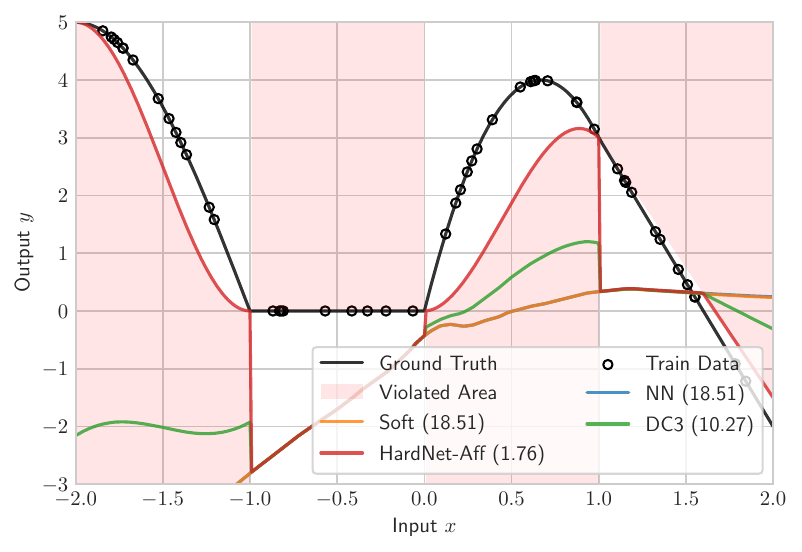}
    \includegraphics[width=0.49\linewidth]{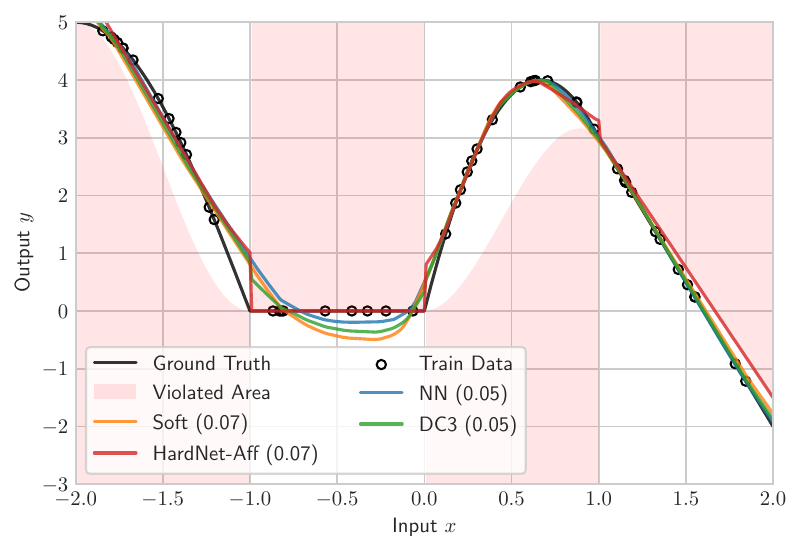}
    \caption{Learned functions at the initial (left) and final (right) epochs with the piecewise constraints. The models are trained on the samples indicated with circles, with their MSE distances from the true function shown in parentheses. \methodaff{} adheres to the constraints from the start of the training. On the other hand, the baselines violate the constraints throughout the training.}
    \label{fig:exp_pwcfull}
\end{figure}

The results in Section~\ref{section:exp_pwc} show \methodaff{} can help generalization on unseen regimes by enforcing constraints. In this section, we provide additional results that train the models on data spanning the entire domain of interest $[-2, 2]$. As shown in Figure~\ref{fig:exp_pwcfull} and Table~\ref{tab:pwcfull}, the models exhibit similar generalization performances while \methodaff{} satisfies the constraints throughout the training.

\begin{table}[t]
    \centering
    \caption{Results for learning with the piecewise constraints. \methodaff{} attains a comparable MSE distance from the true function as other methods without any constraint violation. The max and mean of constraint violations are computed over 401 test samples. Better (resp. worse) values are colored greener (resp. redder). Standard deviations over 5 runs are shown in parentheses.}
    \label{tab:pwcfull}
    \small
    \setlength{\tabcolsep}{3pt}
    \begin{tabular}{lccccc}
        \toprule
        {} & MSE & max violation & mean violation & $T_\textup{test}$ (ms) & $T_\textup{train}$ (s)\\
        \midrule
        NN & \cellcolor[RGB]{222,240,223} 0.04 (0.01) & 
        \cellcolor[RGB]{253,228,218} 0.73 (0.07) & \cellcolor[RGB]{254,254,240} 0.02 (0.01) & \cellcolor[RGB]{205,227,216} 0.14 (0.00) & \cellcolor[RGB]{205,227,216} 2.07 (0.11)\\
        Soft & \cellcolor[RGB]{222,240,223} 0.05 (0.01) &
        \cellcolor[RGB]{253,228,218} 0.69 (0.04) & \cellcolor[RGB]{254,253,239} 0.02 (0.00) & \cellcolor[RGB]{204,224,215} 0.15 (0.01) & \cellcolor[RGB]{222,240,223} 4.00 (0.07)\\
        DC3  & \cellcolor[RGB]{222,240,223} 0.04 (0.01) &
        \cellcolor[RGB]{253,228,218} 0.48 (0.03) & \cellcolor[RGB]{250,253,236} 0.01 (0.00) & \cellcolor[RGB]{238,205,211} 6.05 (0.05) & \cellcolor[RGB]{238,205,211} 15.09 (0.68)\\
        \methodaff{}  & \cellcolor[RGB]{222,240,223} 0.06 (0.01) & \cellcolor[RGB]{208,232,219} 
        0.00 (0.00) & \cellcolor[RGB]{208,232,219} 0.00 (0.00) & \cellcolor[RGB]{218,238,222} 0.86 (0.00) & \cellcolor[RGB]{243,250,229} 5.50 (0.18)\\
        \bottomrule
    \end{tabular}
\end{table}

\subsection{Additional Experiment: Learning with Piecewise Constraints for Both Upper and Lower Bounds}
\label{appendix:pwcbox}

Building on the piecewise–constraint study in Section~\ref{section:exp_pwc}, we next assess \methodaff{} with more complex constraints in which each regime is governed by both upper and lower bounds (or, in the degenerate case, an equality).

We adopt the following ground-truth function:
\begin{equation*}
    f(x) \!=\! \begin{cases}
        -5\sin{\frac{\pi}{2}(x\!+\!1)}-2
        & \textup{if } x\leq -1\\
        -2 & \textup{if } x\in(-1,0]\\
        2-9(x-\frac{2}{3})^2
        & \textup{if } x\in(0,1]\\
        \frac{3}{x^2}-2 & \textup{if } x>1
    \end{cases},
\end{equation*}
subject to the following piecewise box constraints:
\begin{equation*}
    \textup{Constraints}\!:\! \begin{cases}
        5\sin^2{\frac{\pi}{2}(x+1)}-2 \leq f(x) \leq -3\sin{\frac{\pi}{2}(x\!+\!1)}
        & \textup{if } x\leq -1\\
        f(x) = -2 & \textup{if } x\in(-1,0]\\
        \big(4\!-\!9(x\!-\!\frac{2}{3})^2\big)x \leq f(x) \leq 3-4(x-0.5)^2
        & \textup{if } x\in(0,1]\\
        \frac{3}{x^3}-2 \leq f(x) \leq 2 & \textup{if } x>1
    \end{cases}.
\end{equation*}

These piecewise constraints can be aggregated compactly into a boxed input-dependent affine constraint
\(
b^{\ell}(x)\le a(x)f(x)\le b^{u}(x)
\)
with
\begin{equation*}
a(x)= 1,\qquad
(b^{\ell}(x),\,b^{u}(x))=
\begin{cases}
\bigl(5\sin^{2}\!\tfrac{\pi}{2}(x+1)-2,\;
      -3\sin\!\tfrac{\pi}{2}(x+1)\bigr), & x \le -1,\\[2pt]
(-2,\,-2),                               & x \in (-1,0],\\[2pt]
\bigl((4-9(x-\tfrac{2}{3})^{2})\,x,\;
      3-4(x-0.5)^{2}\bigr),              & x \in (0,1],\\[2pt]
\bigl(\tfrac{3}{x^{3}}-2,\;2\bigr),      & x > 1.
\end{cases}
\end{equation*}

\begin{figure}[t]
    \centering
    \includegraphics[width=0.49\linewidth]{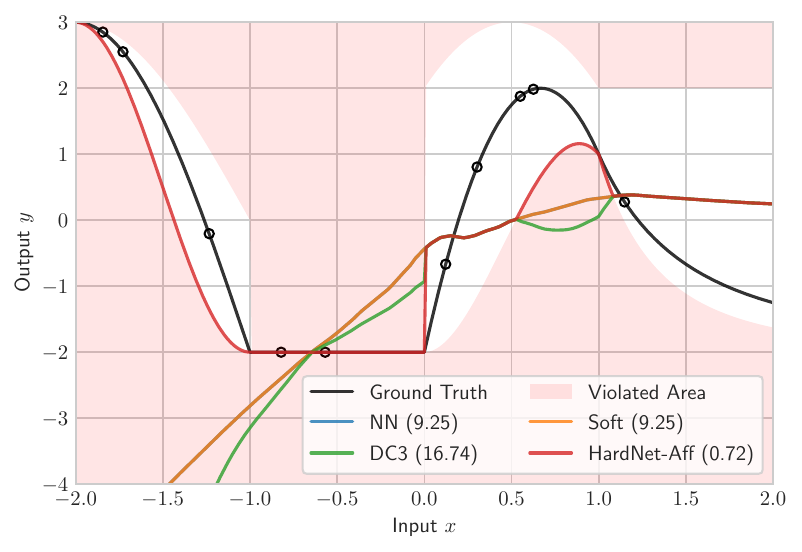}
    \includegraphics[width=0.49\linewidth]{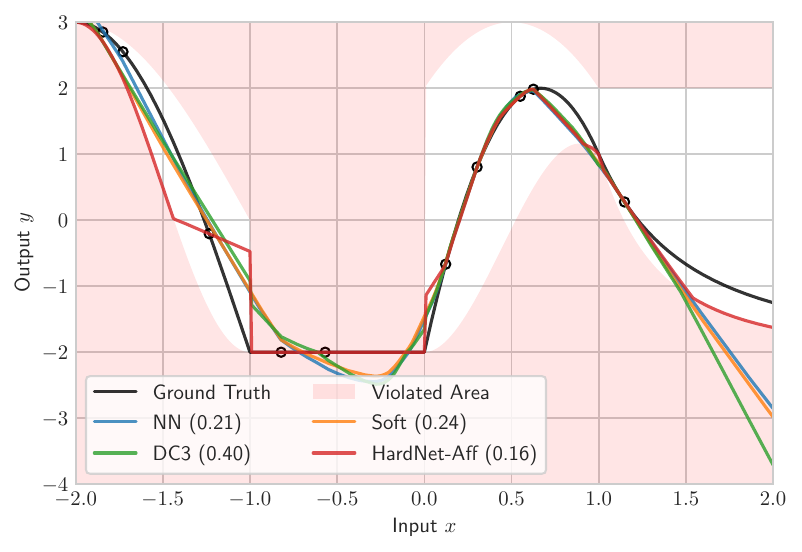}
    \caption{Learned functions at the initial (left) and final (right) epochs with the piecewise box constraints. The models are trained on the samples indicated with circles, with their MSE distances from the true function shown in parentheses. \methodaff{} adheres to the constraints from the start of the training. On the other hand, the baselines violate the constraints throughout the training.}
    \label{fig:exp_pwcbox}
\end{figure}

In this experiment, all models are trained on ten randomly sampled points in the domain $[-2,2]$. 
Figure~\ref{fig:exp_pwcbox} contrasts the learned functions at epoch~0 and epoch~1000. Despite this extreme sparsity, \methodaff{} satisfies the box constraints from the very first epoch and retains feasibility throughout training. While the baselines breach the constraints at many inputs, \methodaff{} respects them everywhere and typically tracks the true function more closely in regions with no data. Quantitative metrics averaged over five runs appear in Table~\ref{tab:pwcbox}. \methodaff{} attains competitive mean-squared error (MSE) while incurring \emph{zero} violation, at the cost of only a modest increase in training time.

\begin{table}[t]
    \centering
    \caption{Results for learning with the piecewise constraints. \methodaff{} attains a comparable MSE distance from the true function as other methods without any constraint violation. The max and mean of constraint violations are computed over 401 test samples. Better (resp. worse) values are colored greener (resp. redder). Standard deviations over 5 runs are shown in parentheses.}
    \label{tab:pwcbox}
    \small
    \setlength{\tabcolsep}{3pt}
    \begin{tabular}{lccccc}
        \toprule
        {} & MSE & max violation & mean violation & $T_\textup{test}$ (ms) & $T_\textup{train}$ (s)\\
        \midrule
        NN & \cellcolor[RGB]{222,240,223} 0.14 (0.04) & 
        \cellcolor[RGB]{246,213,211} 0.92 (0.15) & \cellcolor[RGB]{254,254,240} 0.12 (0.02) & \cellcolor[RGB]{205,227,216} 0.14 (0.00) & \cellcolor[RGB]{205,227,216} 2.52 (0.09)\\
        Soft & \cellcolor[RGB]{222,240,223} 0.15 (0.05) &
        \cellcolor[RGB]{246,213,211} 0.99 (0.18) & \cellcolor[RGB]{254,253,239} 0.11 (0.02) & \cellcolor[RGB]{204,224,215} 0.16 (0.03) & \cellcolor[RGB]{222,240,223} 2.56 (0.07)\\
        DC3  & \cellcolor[RGB]{250,253,236} 0.20 (0.10) &
        \cellcolor[RGB]{238,205,211} 1.27 (0.43) & \cellcolor[RGB]{250,253,236} 0.12 (0.05) & \cellcolor[RGB]{238,205,211} 6.84 (0.04) & \cellcolor[RGB]{238,205,211} 16.61 (0.16)\\
        \methodaff{}  & \cellcolor[RGB]{222,240,223} 0.15 (0.01) & \cellcolor[RGB]{208,232,219} 
        0.00 (0.00) & \cellcolor[RGB]{208,232,219} 0.00 (0.00) & \cellcolor[RGB]{218,238,222} 0.93 (0.01) & \cellcolor[RGB]{243,250,229} 5.29 (0.15)\\
        \bottomrule
    \end{tabular}
\end{table}

\subsection{Details for the Safe Control Experiment}
\label{appendix:cbf}

In this experiment, we consider controlling a unicycle system with system state $x=[x_p, y_p, \theta, v, w]^\top$ which represents the pose, linear velocity, and angular velocity. The dynamics of the unicycle system is given by
\begin{equation*}
    \begin{bmatrix}
        \dot{x_p}\\\dot{y_p}\\\dot{\theta}\\
        \dot{v}\\\dot{w}
    \end{bmatrix}
    =
    \begin{bmatrix}
        v\cos{\theta}\\
        v\sin{\theta}\\
        w\\0\\0
    \end{bmatrix}
    +
    \begin{bmatrix}
        0 & 0\\
        0 & 0\\
        0 & 0\\
        1 & 0\\
        0 & 1\\
    \end{bmatrix}
    \begin{bmatrix}
        a_\textup{lin}\\ a_\textup{ang}
    \end{bmatrix},
\end{equation*}
with the linear and angular accelerations $a_\textup{lin}, a_\textup{ang}$ as the control inputs.

To avoid an elliptical obstacle centered at $(c_x, c_y)$ with its radii $r_x, r_y$, one could consider the following CBF candidate:
\begin{equation*}
    h_\textup{ellipse}(x) = 
    \Big(\dfrac{c_x - (x_p + l\cos{\theta})}{r_x}\Big)^2
    + \Big(\dfrac{c_y - (y_p + l\sin{\theta})}{r_y}\Big)^2 -1,
\end{equation*}
where $l$ is the distance of the body center from the differential drive axis of the unicycle system. However, it is not a valid CBF since the safety condition~\eqref{eq:cbf_constraint} does not depend on the control input (i.e., $\nabla h_\textup{ellipse}(x)^\top g(x)=0 \;\;\forall x$). Instead, we can exploit a higher-order CBF (HOCBF) given by
\begin{equation*}
    h(x) = \dot{h}_\textup{ellipse}(x) + \kappa h_\textup{ellipse}(x),
\end{equation*}
for some $\kappa>0$. Then, ensuring $h\geq 0$ implies $h\geq0$ given $h(x(0))\geq0$, and the safety condition~\eqref{eq:cbf_constraint} for this $h$ depends on both control inputs $a_\textup{lin}, a_\textup{ang}$. Refer to~\citet{tayal2024collision} for a detailed explanation.

The goal of this problem is to optimize the neural network policy $\pi_\theta(x)=\pi_\textup{nom}(x)+f_\theta(x)$ to minimize the expected cost over the trajectories from random initial points within the range from $[-4, 0, -\pi/4, 0, 0]$ to $[-3.5, 0.5, -\pi/8, 0, 0]$.
For an initial state sample $x_s$, we consider the cost of the rolled-out trajectory through discretization with time step $\Delta t = 0.02$ and $n_\textup{step}=50$ as
\begin{equation}\label{eq:traj_cost}
    \Delta t \sum_{i=0}^{n_\textup{step}-1} x_i^\top Q x_i + \pi_\theta(x_i)^\top R \pi_\theta(x_i),
\end{equation}
where $x_i$ is the state after $i$ steps, and $Q = diag(100, 100, 0, 0.1, 0.1)$ and $R = diag(0.1, 0.1)$ are diagonal matrices. The neural network policies are optimized to reduce~\eqref{eq:traj_cost} summed over 1000 randomly sampled initial points.

\end{document}